\documentclass[5p,times]{elsarticle}
\usepackage[utf8x]{inputenc}
\usepackage[T1]{fontenc}
\usepackage{tikz}
\usetikzlibrary{shadings}
\usetikzlibrary{spy,calc}   
\usepackage{pgfplots}
\pgfplotsset{compat=newest}
\usepackage[raggedright]{subfigure}
\usepackage{amssymb}
\usepackage{float}

\usepackage{array}
\usepackage{amsthm}
\usepackage[intlimits]{amsmath}
\usepackage{amsmath}
\usepackage{booktabs}
\usepackage{multirow}
\usepackage[round]{natbib}
\biboptions{round,authoryear}
\usepackage{xcolor}
\usepackage{orcidlink}
\usepackage[hyphenbreaks]{breakurl}

\usepackage{makecell}
\usepackage{tabularx} 

\definecolor{Green}{RGB}{85.4, 140.3, 130.5}

\journal{Expert Systems with Applications}

\begin{document}

\begin{frontmatter}



\title{Stylometry recognizes human and LLM-generated texts in short samples}


\author[1,3]{Karol Przystalski \orcidlink{0000-0002-8572-1469}}
\ead{karol.przystalski@uj.edu.pl}

\author[2,3]{Jan K. Argasiński \orcidlink{0000-0002-2992-718X}}
\ead{jan.argasinski@uj.edu.pl}

\author[3]{Iwona Grabska-Gradzińska \orcidlink{0000-0002-5799-5438}} 
\ead{iwona.grabska@uj.edu.pl}

\author[3,4]{Jeremi K. Ochab \orcidlink{0000-0002-7281-1852}}
\ead{jeremi.ochab@uj.edu.pl}


\address[1]{Exadel, Na Zjeździe 11, 30-527 Kraków, Poland}

\address[2]{Sano -- Centre for Computational Medicine, Czarnowiejska 36/C5, 30-054 Kraków}

\address[3]{Faculty of Physics, Astronomy and Applied Computer Science, Jagiellonian University, Łojasiewicza 11, 30-348 Kraków, Poland.}

\address[4]{Mark Kac Centre for Complex Systems Research, Jagiellonian University, Łojasiewicza 11, 30-348 Kraków, Poland.}




\begin{abstract}
The paper explores stylometry as a method to distinguish between texts created by Large Language Models (LLMs) and humans, addressing issues of model attribution, intellectual property, and ethical AI use. Stylometry has been used extensively to characterise the style and attribute authorship of texts. By applying it to LLM-generated texts, we identify their emergent writing patterns.
The paper involves creating a benchmark dataset based on Wikipedia, with (a) human-written term summaries, (b) texts generated purely by LLMs (GPT-3.5/4, LLaMa 2/3, Orca, and Falcon), (c) processed through multiple text summarisation methods (T5, BART, Gensim, and Sumy), and (d) rephrasing methods (Dipper, T5).
The 10-sentence long texts were classified by tree-based models (decision trees and LightGBM) using human-designed (StyloMetrix) and n-gram-based (our own pipeline) stylometric features that encode lexical, grammatical, syntactic, and punctuation patterns. 
The cross-validated results reached a performance of up to .87 Matthews correlation coefficient in the multiclass scenario with 7 classes, and accuracy between .79 and 1. in binary classification, with the particular example of Wikipedia and GPT-4 reaching up to .98 accuracy on a balanced dataset.
Shapley Additive Explanations pinpointed features characteristic of the encyclopaedic text type, individual overused words, as well as a greater grammatical standardisation of LLMs with respect to human-written texts.
These results show -- crucially, in the context of the increasingly sophisticated LLMs -- that it is possible to distinguish machine- from human-generated texts at least for a well-defined text type.


\end{abstract}




\end{frontmatter}



 \section{Introduction }
 \label{sec:Introduction}

 In the dynamicaly expanding field of natural language processing, Large Language Models (LLMs), introduced by transformative models like GPT, have revolutionized approaches to analyze language by enabling machines to mimic human-like text generation. As the use of pretrained AI models becomes increasingly common, growing concerns around issues like ownership, attribution, intellectual property rights, and responsible usage highlight the urgent need for advanced methods to ensure ethical deployment and proper crediting of AI-generated work -- alongside the development of reliable model detection tools.

The problem of stylometry and authorship attribution is a crucial aspect in this context. Stylometry, meaning the quantitative study of linguistic style patterns, is a valuable tool for effective text differentiation. By examining subtle differences in writing style, one can discover unique markers that distinguish one author from another. Stylometric features provide a detailed understanding of the characteristics of particular LLMs, offering a granular approach towards model identification. This not only facilitates differentiation but also enhances our comprehension of the linguistic idiosyncrasies ingrained in these models. The challenge lies in accurately attributing text to the correct author or model, especially as language models grow more sophisticated and their outputs increasingly indistinguishable from human writing.
This paper explores the utilization of machine learning techniques in identifying  stylistic markers and patterns that are characteristic for specific language models, augmenting our ability to differentiate them with greater accuracy. By focusing on properties such as the word choice and syntactic patterns, our aim is to uncover the linguistic fingerprints that distinguish one model results from another.



The exploration of stylometry in model detection and differentiation reaches far beyond technical considerations towards ethical implications. Understanding the characteristic stylometric properties of language models productions contributes to securing the responsible AI practices, promoting transparency and accountability. LLM safety and ethics are the most important concerns in this regard. Ensuring that language models are used ethically should account for such issues as bias, misinformation, and the potential for generating harmful content.
By promoting stylometry, this paper aims to provide a distinctive perspective
,thereby contributing to a more comprehensive understanding of language model deployment in diverse applications. This approach not only improves our ability to safeguard intellectual property but also cultivates a culture of responsibility and trust in the AI community.

The research presented in this paper provides an innovative approach to distinguish between models. 
As we navigate the complex interplay of technology, ethics, and stylometry, our goal is to contribute to the responsible advancement of natural language processing technologies.

The main contributions of this paper are as follows:
\begin{enumerate}
\item \textbf{Application of Stylometry to Differentiate Texts}: The paper applies stylometry to distinguish between texts generated by LLMs and human-authored texts. Stylometry, traditionally used for authorship attribution and literary style analysis, is shown to be effective in identifying writing patterns specific to LLMs.

\item \textbf{Creation of a Diverse Dataset}: The study constructs a dataset based on (a) human-written Wikipedia texts, (b) their summaries processed through various text summarization methods (T5, BART, Gensim, and Sumy), and (c) summaries generated by LLMs (GPT-3.5, GPT-4, LLaMa~2/3, Orca, and Falcon) prompted a given term only. This dataset allows for a comprehensive analysis of different text generation methods.


\item \textbf{High Classification Performance}: The study demonstrates that tree-based classifiers (decision trees and LightGBM) can achieve high performance in classifying texts, reaching up to 0.87 Matthews correlation coefficient in multiclass scenarios (with 7 classes) and up to 1.00 accuracy in binary classification (e.g., distinguishing Wikipedia from GPT-4-generated texts at 0.98 accuracy).

\item \textbf{Insights into LLM and Human Text Characteristics}: The paper provides detailed insights into specific features that differentiate LLM-generated texts from human-authored texts. It highlights that LLM-generated texts tend to have more grammatical standardization and may overuse certain words or punctuation marks compared to human-written texts.

\item \textbf{Implications for Ethical AI Use}: The paper emphasizes the need for robust methods to track and identify AI-generated outputs to ensure ethical AI use, addressing concerns around model attribution, intellectual property, and responsible deployment of AI technologies.

\item \textbf{Potential for Stylometry in Future AI Applications}: The research suggests that stylometry could continue to be a valuable tool for distinguishing machine-generated texts from human-authored ones, especially as LLMs become more sophisticated, highlighting its potential role in future AI applications and governance.
\end{enumerate}

This manuscript is structured into six sections including: 1.~\nameref{sec:Introduction}, 2.~\nameref{sec:Related}, 3.~\nameref{sec:Methodology}, 4.~\nameref{sec:Results}, 5.~\nameref{sec:Discussion}, and 6.~\nameref{sec:further}. 

In the \nameref{sec:Introduction} the rationale for the presented research is provided. In the \nameref{sec:Related} we present important background for our work. The design of our own experiments is detailed in \nameref{sec:Methodology}. \nameref{sec:Results} of the classification are visualised in the next section. Finally the \nameref{sec:Discussion} and \nameref{sec:further} section include general remarks, known limitations and possible future directions for the research along with the inventory of crucial findings.

\section{Related works}
\label{sec:Related}

Stylometry, the study of linguistic style, has long been a important tool in authorship attribution, and its relevance has grown significantly with the advent of Large Language Models.
As these models produce increasingly human-like text, the ability to distinguish between human-authored and machine-generated texts (MGT) becomes essential, not just for academic and forensic purposes, but also for ensuring the safety and ethical use of LLMs.

In this section we present works relevant to the theme of stylometry itself and related to MGT, particularly by LLMs; we mention research about stylometric modeling; and finally showcase papers that tackle the theme of safety and ethics regarding emerging generative linguistic tools. 

\subsection{Stylometry and author attribution}

\cite{neal2017surveying} in \textit{Surveying Stylometry Techniques and Applications} provide an extensive overview of stylometry research, focusing on authorship attribution, verification, profiling, stylochronometry, and adversarial stylometry. The survey is in depth, covering different subtasks, datasets, experimental methods, and contemporary approaches. It includes detailed performance analysis taking into account 1,000 authors using 14 different algorithms. The paper exposes key challenges such as scaling authorship analysis to account for a large number of authors with minimal text samples available. It also presents ongoing research challenges and showcases different software tools that support stylometric analysis - both open-source and commercial options. 

\textit{A survey of modern authorship attribution methods} \citep{stamatatos2009survey} gives an detailed presentaion of the various computational methods utilized in the field of authorship attribution. It traces the evolution of these methods from their inception in the 19th century, highlighted by the seminal study of \cite{mosteller1968association}, to the contemporary techniques that leverage statistical and computational approaches. This survey discusses the main characteristics, strengths, and weaknesses of modern authorship attribution methods.

\subsection{Stylometric modeling}

Paper titled \textit{TDRLM: Stylometric learning for authorship verification by Topic-Debiasing} \citep{hu2023tdrlm} proposes a “Topic-Debiasing Representation Learning Model” (TDRLM) to enhance stylometric authorship verification. The TDRLM utilizes a topic-debiasing attention mechanism with position-specific topic scores to mitigate the influence of topical bias in tokenized texts. Experimental results demonstrate that the TDRLM outperforms current state-of-the-art stylometric learning models and advanced language models, achieving the highest Area Under Curve (AUC) scores of 92.47\% for the Twitter-Foursquare dataset and 93.11\% for the ICWSM Twitter dataset. The study highlights that topic-related words can negatively impact machine learning algorithms for authorship verification, prompting the development of the TDRLM model to improve verification accuracy.

The evolution of current methods is well exemplified by a series of papers by \cite{kumarage_stylometric_2023,kumarage2023neural,bhattacharjee_conda_2023}.
\cite{kumarage_stylometric_2023} and \cite{kumarage2023neural} used a fusion architecture of fine-tuned RoBERTa augmented with a combination of stylometric features -- lexical, syntactic, and structural such as lexical richness, readability, punctuation counts, word / sentence / paragraph counts etc.
Interestingly, the authors used as a baseline XGBoost with either stylometric or bag-of-word features, which allowed them to use SHAP explanation in the same vein as we do in the present paper.
The fusion was proved beneficial especially for short texts (Twitter timelines) and limited training data, but out-of-distribution problems (cross-domain or unseen LLMs) remained challenging.
To improve these issues, \cite{bhattacharjee_conda_2023} turned away from stylometric classfiers to self-supervised contrastive learning and unsupervised domain adaptation techniques at the cost of losing the explainability.



\subsection{Authorship-stylometry and LLMs}

\textit{Large Language Models: A Survey} by \cite{zhao2023survey} provides a comprehensive overview of LLMs, their development, capabilities, and applications. The authors review notable LLMs, such as GPT, LLaMa, and PaLM, discussing their design, strengths, and limitations. The paper explores various methods used for constructing and enhancing LLMs, examines key datasets utilized for training and evaluation, and assesses these models' performance across standard benchmarks. It highlights LLMs' significant advancements in natural language tasks, largely attributable to their training on massive datasets, reflecting the importance of data scale in model performance.

\cite{argamon2018computational} contributes with \textit{Computational Forensic Authorship Analysis: Promises and Pitfalls} -- a comprehensive examination of the techniques involved in computational authorship analysis, focusing on their application within legal and forensic contexts. Authors highlight how these methods have advanced to the point of being reliable enough for real-world legal applications, underscoring their evolution and growing acceptance in rigorous environments. Paper discusses various computational methods, detailing their underlying assumptions, necessary analytic controls, and the crucial reliability testing they must undergo to ensure their effectiveness. Moreover, the paper addresses the potential pitfalls of these techniques, offering guidance to practitioners on how to achieve results that are not only trustworthy but also comprehensible. 

\textit{Learning Stylometric Representations for Authorship Analysis} \citep{ding2017learning} explores a neural network approach to learn stylometric representations that capture various linguistic features such as topical, lexical, syntactical, and character-level characteristics. This methodology aims to improve the tasks of authorship characterization, identification, and verification by mimicking the human sentence composition process and incorporating these diverse linguistic categories into a distributed representation of words. The effectiveness of this approach is demonstrated through extensive evaluations across multiple datasets, including Twitter, blogs, reviews, novels, and essays, where the proposed models notably outperform traditional stylometric and other baseline methods. This research highlights the potential of neural networks in extracting and utilizing complex stylistic features for detailed authorship analysis in diverse textual domains.

With the question \textit{Can Large Language Models Identify Authorship?} \cite{huang2024can} explores the capabilities of LLMs in performing authorship verification and attribution tasks without requiring domain-specific fine-tuning. The authors demonstrate that LLMs can effectively conduct zero-shot, end-to-end authorship verification and accurately attribute authorship among multiple candidates. Furthermore, the study sift how these models can offer explainability in their analysis, focusing particularly on the role of linguistic features. 

\textit{Learning Interpretable Style Embeddings via Prompting LLMs} \citep{patel2023learning} presents an innovative approach for deriving interpretable style embeddings, called LISA embeddings, from LLMs using prompting techniques. The authors address the challenge of uninterpretable style vectors commonly produced by current neural methods in style representation learning, which are problematic for tasks that require high interpretability like authorship attribution. To overcome this, they employ prompting to generate a synthetic dataset of stylometric annotations. This dataset facilitates the training of LISA embeddings, which are designed to be interpretable and useful for analyzing author styles in texts. Additionally, the authors contributed by releasing both the synthetic stylometry dataset and the LISA style models, enabling further exploration and development in the field of stylometry and style analysis.

\textit{A model-independent redundancy measure for human versus ChatGPT authorship discrimination using a Bayesian probabilistic approach} \citep{bozza2023model} introduces a novel method to distinguish between human-authored texts and those generated by AI models like ChatGPT. This approach utilizes a model-independent redundancy measure that effectively captures syntactical differences between human and machine-generated texts. The researchers employed a Bayesian probabilistic framework, specifically using the Bayes factor, to provide a robust and consistent classification criterion. This method proves particularly effective even with short text samples, demonstrating its potential utility in forensic and other analytical settings where distinguishing between human and AI authorship is crucial. The study highlights the applicability of this technique across various languages and text genres, indicating its broad potential for addressing the challenges posed by the increasing sophistication of MGT in academic and professional contexts.

Authors of \textit{Who Wrote it and Why? Prompting Large Language Models for Authorship Verification} \citep{hung2023wrote} offer a new technique named \texttt{PromptAV}. This method utilizes Large Language Models (LLMs) to perform authorship verification effectively and with improved interpretability. Authors claim that the \texttt{PromptAV}, demonstrates improved performance compared to existing state-of-the-art baselines, particularly in scenarios with limited training data. It enhances interpretability by providing intuitive explanations, making it a promising tool for applications in forensic analysis, plagiarism detection, and identifying deceptive content in texts. This approach is meant to address the current limitations of traditional stylometric and deep learning methods, which typically require extensive data and lack explainability (e.g., \citealp{bhattacharjee_conda_2023}).

The paper \textit{T5 meets Tybalt: Author Attribution in Early Modern English Drama Using Large Language Models} \citep{hicke2023t5} explores the application of LLMs for authorship identification in Early Modern English drama. The study finds that LLMs, specifically a fine-tuned T5-large model, can accurately predict the author of short passages and outperform traditional baselines like logistic regression, SVM with a linear kernel, and cosine delta. However, the presence of certain authors in the model's pretraining data introduces biases, leading to occasional confident misattributions of texts. This highlights both the promising potential and the concerning limitations of using LLMs for stylometric analysis in literary studies.

Finally, the paper titled \textit{Detecting ChatGPT: A Survey of the State of Detecting ChatGPT-Generated Text} \citep{dhaini2023detecting} provides an overview of current approaches for identifying text generated by ChatGPT. It highlights the challenges of distinguishing between human-written and machine-generated content, especially given the high fluency and human-like quality of ChatGPT outputs. The survey reviews various datasets specifically created for this detection task, examines different methodologies employed, and discusses qualitative analyses that help identify characteristics unique to ChatGPT-generated text. It also explores the broader implications for domains such as education, law, and science, emphasizing the need for effective MGT detection methods to maintain content integrity.

\subsection{LLM detection benchmarks}
A non-exhaustive list of datasets designed for MGT detection can be found in \cite{wu_survey_2025}. 
Most of these benchmark datasets are English-only (TuringBench \cite{uchendu_turingbench_2021}, CHEAT \cite{yu_cheat_2025}, OpenLLMText \cite{chen_token_2023}, GROVER \cite{zellers_defending_2019}, TweepFake \cite{fagni_tweepfake_2021}, ArguGPT \cite{liu_argugpt_2023}, MAGE \cite{li_mage_2024}, PAN's Voight-Kampff Generative AI Detection task \cite{bevendorff_overview_2024}). Others include Spanish (AuTexTification \cite{sarvazyan_overview_2023}), Chinese (HC3, HC3 Plus \cite{guo_how_2023, su_hc3_2024}) or rarely they are multilingual (MULTITuDE \cite{macko_multitude_2023}, M4 \cite{wang_m4_2024}).
The more recent ones are also multi-domain (\cite{macko_multitude_2023, wang_m4_2024, bevendorff_overview_2024, li_mage_2024, sarvazyan_overview_2023, li_mage_2024}), and use diverse LLM generators (see especially TuringBench \cite{uchendu_turingbench_2021}, MAGE \cite{li_mage_2024}), which is particularly challenging to collect for multiple languages. Such multi-generator and multi-domain benchmarks allow one to test generalizability of the classifiers to unseen domains and unseen LLMs, a realistic scenario considering the rate of development of both closed and open-source LLMs.
An even more comprehensive overview in terms of domains and languages is given in \cite{macko_multitude_2023}.
It is worth stressing that collecting and generating a well-controlled benchmark with multiple domains, generators, and languages is a considerable endeavour, as is well-known in corpus linguistics~\cite{ludeling_corpus_2008}.

When evaluating MGT detectors, caution is required with regard to training on data external to these benchmarks, as some of the data were collected from other primary sources.

\subsection{LLM detection methods}
We refer to the \textit{Overview of the “Voight-Kampff” Generative AI Authorship Verification Task at PAN and ELOQUENT 2024} by \cite{bevendorff_overview_2024} as a recent benchmark of available methods.
The submissions included mainly (i) perplexity-based systems, (ii) term-based systems (iii) and ensembles of both.
The first ones rely mainly on the perplexities of a set of known LLMs (which is a limitation on its own) used as features for a classifier. The second class uses fine-tuned classifiers (often neural ones such as modified versions of BERT or their ensembles) with word embeddings or linguistic (stylometric) features; some other involve fine-tuned generative LLMs (regarded as unreliable by \citealp{wu_survey_2025}).
Several proposals, e.g., \cite{guo_machine-generated_2024,miralles_team_2024,yadagiri_team_2024,guo_blgav_2024}, made advantage of various sets stylometric features, while others found it useful to augment data or simply expand the training dataset.
A noteworthy example is \cite{lorenz_baselineavengers_2024}, whose SVM classifier based on tf-idf features (sic!) by was ranked third beating all neural baselines and most of the neural-based competitors, mostly thanks to its robustness to the many obfuscation strategies designed by the Task’s authors.
\textit{We regard our boosted trees classifier to follow a similar simplistic, inexpensive but effective design, often overlooked in recent research.}

Among many other methods reviewed by \cite{wu_survey_2025} or \cite{crothers_machine-generated_2023} that were not represented in the PAN's task are the logits-based statistics. This family encompasses primarily zero-shot methods which, however, require access to a surrogate (often weaker) language model or, ideally, to the source LLM -- hence they are dubbed white-box methods -- to obtain its raw outputs that in turn allow to determine the likelihood of a text being generated by the LLM.
The black-box statistical methods, on the other hand, do not require such access. Instead, given an original text they machine-regenerate it and subsequently compare these versions to obtain a similarity score.

Lastly, a whole strain of research has been MGT watermarking, wherein an imperceptible signature is embedded in the generated texts either by insertion of modified training samples (to defend against unauthorised LLM fine-tuning), manipulating the logits output distribution or token sampling, or character- and word-level replacements at the post-processing stage (especially useful when using black-box models). 
\cite{sadasivan_can_2025,lu_large_2024} report successful attack strategies on some of such watermarking schemes by iterative paraphrasing or probing a watermarked LLM to infer the signatures.
\textit{In the present paper, we assume that the adversary does not use a watermarked LLM}.

\subsection{Robustness of LLM detection}


A number of issues can degrade the performance of MGT detection. \cite{wu_survey_2025} divided them into out-of-distribution challenges and attacks.

The first type includes detection across domains (i.e., usually different text types or genres involving different vocabulary, style, topics, and overall distribution of textual features), across languages (but also including texts written by non-native speakers), and across LLMs (i.e., generally detection of LLMs not available during the detector's training).
In the latter case, there are some reports~\citep{sarvazyan_supervised_2023} that supervised MGT detectors tend to generalise well across LLM scales but less so across their model families. For neural detectors, therefore, incorporating
MGT from various sources is recommended, since an additional fine-tuning -- even on small samples -- can effectively alleviate this issue.
\textit{In our study, no data external to the dataset described in Sec.~\ref{sec:Dataset} is used for training, and we do not tackle the issue of cross-language detection} (reported as challenging by \citealp{bevendorff_overview_2024}).
\textit{The cross-domain detection is tested on an existing benchmark by \cite{sarvazyan_overview_2023}.}

The potential attacks include: paraphrase (where LLM output is subsequently paraphrased by another model in order to change the textual feature distribution of the original MGT; \citealp[see, e.g.,]{sadasivan_can_2025}), adversarial (involving textual perturbations on the level of characters like various misspelling strategies,  \citealp[see ]{stiff_detecting_2022}, syntax, \citealp[see ]{bhat_how_2020},  or lexis, \citealp[see ]{crothers_adversarial_2022}), prompt (using complex and varied prompts for MGT, see \citealp{guo_how_2023,liu_detectability_2024}) attacks and models trained specifically to confound existing detectors.
These attacks affect MGT detectors differently, depending on whether they are watermarking-based (specifically targeted by paraphrase and adversarial attacks), zero-shot or fine-tuned supervised detectors.
The latter can effectively defend against some of these attacks by continually expanding training datasets, e.g., with adversarial examples.
Notably, \cite{bevendorff_overview_2024} report that in the joint PAN and ELOQUENT detection-obfuscation task, none of the obfuscation submissions managed to beat in terms of their difficulty simple methods such as Unicode obsfuscations or shortening text length.
\textit{In this study, we perform a one-step paraphrase attack (repeated paraphrasing is possible, as in \citealp{sadasivan_can_2025}), but we do not include any attacks in the training data.}

We do not cover the issue of mixed texts (human-edited MGT or LLM-edited human-written texts or texts whose separate parts come from either human or machine).





\subsection{LLMs safety and ethics}

The application of stylometry to LLMs is particularly important given the potential risks associated with their misuse, such as the generation of misleading information, deepfake text, or malicious content, as described below.
We note, however, that mere detection that a text has been machine-generated -- which is the objective of the present paper -- does not imply that it is untrustworthy or malicious.
\cite{schuster_limitations_2020} reported that, even though human language tends to stylistically change when deceiving, stylometry fails to detect malicious use of (now perhaps obsolete) LLMs, and that such issues involve a whole ecosystem of fraud (including among others fact-checking, users' feedback, and content propagation through social networks).

\textit{A Survey of Safety and Trustworthiness of Large Language Models through the Lens of Verification and Validation} \citep{huang2024survey} provides a detailed examination of the safety and trustworthiness concerns associated with LLMs. It categorizes the known vulnerabilities of LLMs into three main types: inherent issues, external attacks, and unintended bugs. The study extends traditional verification and validation (V\&V) techniques, commonly used in software and deep learning model development, to enhance the safety and reliability of LLMs throughout their lifecycle. Specifically, the survey discusses four complementary V\&V techniques: falsification and evaluation, verification, runtime monitoring, and the implementation of regulations and ethical guidelines. These approaches are aimed at ensuring that LLMs align with safety and trustworthiness requirements, addressing both existing challenges and potential risks.

Another survey -- \textit{on Large Language Model (LLM) Security and Privacy: The Good, the Bad, and the Ugly} \citep{yao2024survey} offers a detailed exploration of the security and privacy dimensions associated with LLMs. It assesses how LLMs can both enhance and threaten cybersecurity in various applications. The authors categorize their findings into beneficial uses ("The Good"), such as improving code security and data privacy, offensive applications ("The Bad"), like their use in user-level attacks due to their sophisticated reasoning capabilities, and inherent vulnerabilities ("The Ugly") that could be exploited maliciously. The survey emphasises the dual nature of LLMs in cybersecurity, showcasing their potential to advance security measures while also posing significant risks if not carefully managed and regulated. Furthermore, it identifies areas needing further research, such as model and parameter extraction attacks and the development of safe instruction tuning, underlining the complexity and evolving nature of LLM applications in security contexts.

\textit{Adversarial stylometry: Circumventing authorship recognition to preserve privacy and anonymity} \citep{brennan2012adversarial} introduces the field of adversarial stylometry. This research area focuses on strategies like obfuscation and imitation to effectively counter authorship recognition methods, which are crucial for maintaining privacy and anonymity in written communication. The study demonstrates that manual techniques, where individuals intentionally alter their writing style, are particularly effective at evading detection, often reducing the accuracy of stylometric tools to the level of random guesses. Even individuals with no prior knowledge of stylometry or limited time investment can successfully employ these strategies. Additionally, the paper discusses the efficacy of various obfuscation techniques and highlights the limited effectiveness of automated methods such as machine translation. 

\textit{ChatGPT and a new academic reality: Artificial Intelligence‐written research papers and the ethics of the large language models in scholarly publishing} \citep{lund2023chatgpt} addresses the transformative effects of ChatGPT and similar large language models on academic and scholarly environments. Paper highlights several key concerns, including the potential for inherent biases in training data and algorithms that could compromise scientific integrity. Additionally, the it raises critical ethical issues, such as the ownership of content produced by these models and the proper use of third-party content, which are essential for maintaining transparency and fairness in academic publishing. The discussion extends to the responsibilities of researchers and publishers in ensuring that these technologies are utilized in a manner that upholds the ethical standards of scholarly work.

Last, but not least -- \textit{ChatGPT and the rise of large language models: the new AI-driven infodemic threat in public health} by \cite{de2023chatgpt} examines the dual-edged impact of LLMs on public health. It acknowledges the potential of LLMs to aid scientific research through their ability to process and generate large amounts of data quickly. However, it critically highlights the risk of an “AI-driven infodemic”, where the rapid and widespread dissemination of misinformation could be facilitated by these same technologies. The paper calls for urgent policy actions to mitigate these risks, emphasizing the need for a balanced approach in harnessing the benefits of LLMs while safeguarding against their potential to undermine public health and the integrity of scientific research. This includes the establishment of regulatory frameworks and the proactive monitoring of the use of LLMs to prevent the spread of false information.






\section{Metholodogy}
\label{sec:Methodology}

The process of the proposed solution is divided into several steps. The data acquisition and cleaning is explained in the first part of the chapter (\ref{sec:Dataset}). The data was next extended by the summaries generated with various text summarisation methods (\ref{sec:Summarizers}). In the next step, we added additional short terms descriptions generated using different language models (\ref{sec:LanguageModels}). Finally, based on the stylometric features, we differentiate between the texts generated by the models and the humans (\ref{sec:Stylometry}, \ref{sec:Classification}). 

\subsection{Dataset}
\subsubsection{Human texts: Wikipedia summaries}
\label{sec:Dataset}

The dataset is based on Wikipedia terms using two different Python libraries: \texttt{datasets} from HuggingFace\footnote{\url{https://huggingface.co}} \citep{lhoest-etal-2021-datasets} and \texttt{Wikipedia-API}. In the first method we used the dataset from 2022 named \texttt{20220301.simple}.
Similarly to how \cite{bevendorff_overview_2024} collected their data, our choice was dictated by the date of GPT-3.5 release, so that we avoid contaminating the human-authored and edited texts with MGT in the view of its increased presence in Wikipedia \citep{brooks_rise_2024}.
We obtained 1500 terms using the first method and 1048 terms using the second. The final dataset used in this paper consists of 2439 terms. The number is a result of the preprocessing part and the removal of all examples that did not meet one of the following requirements:
\begin{itemize}
    \item the term text consists of at least 1100 alphanumerical characters, including punctuation marks,
    \item consists of at least 10 sentences,
    \item the first 10 sentences do not include references (bibliography).
\end{itemize}
Each term description that did not fulfil the above requirements was removed from the dataset. Before the above validation, non-latin letters were removed, and characters like duplicated whitespaces were removed, including brackets, semicolons, and dots.
For classification purposes, the texts were shortened to a maximum of 18 sentences (this is the maximum number of sentences generated by GPT models, despite the 10-sentence limit).
Additionally, we removed several outlying texts (which had high sentence counts, mainly due to improper spaCy segmentation of lists or enumerations), resulting in 2424 terms.

\subsubsection{Text summarizers}
\label{sec:Summarizers}

We used four text summarisation methods for comparison: (1) A very popular Python method in the gensim library \citep{gensim}. It is already outdated, as there are more complex methods based on transformers that reportedly give better results. (2-3) Transformer-based T5 \citep{t5sum} and BART summarizers \citep{bartsum}. (4) The last summarization method is called sumy and is implemented in the \texttt{sumy}\footnote{\url{https://pypi.org/project/sumy}} library.

Every summarisation method is provided with the Wikipedia terms descriptions, but each has different parameters to be set. We tried to set such parameters to obtain a summary of about 10 sentences for each term. The gensim summarizer does have a `number of sentences' parameter, but we did not set it to an exact number. It produced a sufficient number of sentences and in case it exceeded the limit, we just dropped the excess sentences. For the T5 and BART summarizes we got the best results with setting the maximum number of characters to 1000. The length penalty parameter and number of beans were left to the standard values of 2.0 and 4, respectively. Sumy has a parameter that allows one to set the exact number of sentences, which we set to 10.

Table~\ref{tab:summary_stats} shows basic statistics of the dataset. In particular, T5 tends to produce a highly varying summary length, both in terms of tokens and sentences, and amount of punctuation. The reason is its failure, resulting in repetition of the same letters or words and a number of full stops. The other summarisers do not produce such artefacts, with BART generating a low number of short sentences, Gensim a low number of relatively longer sentences, and Sumy a larger number of sentences. 

\begin{table*}[ht]
\centering
\begin{tabular}{lcccc}
\toprule
    & \textbf{Gensim} & \textbf{Sumy} & \textbf{T5} & \textbf{BART} \\ 
\midrule
\textbf{\#tokens}                      & 71 ± 36     & 249 ± 62  & 220 ± 110 & 61 ± 12   \\
\textbf{fraction of punctuation [\%]} & 13.4 ± 4.9   & 13.8 ± 4.0  & 28 ± 11   & 13.1 ± 4.3 \\
\textbf{\#tokens / sentence}           & 28.8 ± 9.8      & 24.4 ± 6.0    & 14.0 ± 13.0   & 19.7 ± 5.5    \\
\textbf{\#sentences}                   & 2.5 ± 1.0       & 10.4 ± 2.2    & 21.0 ± 15.0   & 3.19 ± 0.7    \\
\textbf{max. \#sentences}              & 9               & 41            & 138           & 7             \\ 
\bottomrule
\end{tabular}
    \caption{Basic summariser dataset statistics: total number of tokens (including punctuation), fraction of punctuation tokens in the total token count, mean number of tokens in a sentence, number of sentences, and the maximal number of sentences. The numbers are averages and standard deviations across all documents.}
    \label{tab:summary_stats}
\end{table*}

\subsubsection{LLM-generated descriptions}
\label{sec:LanguageModels}

Large language models were used to generate term descriptions from scratch, i.e., they were provided only with terms they were prompted to describe, but not with any part of the Wikipedia articles.
We chose six language models, including the open and API-based ones. We used the ChatGPT API for two models: GPT-3.5-turbo, and GPT-4 \citep{gpt4}. LLaMa~2 and 3 with 7 and 8 billion parameters, respectively \citep{llama2}. In this case, we used the Ollama\footnote{\url{https://ollama.com/}} library. For the other two models: Orca \citep{orca} and Falcon \citep{almazrouei2023falcon} we used the GPT4All library \citep{gpt4all}. The models we used had 8 and 11 billion parameters, respectively. We used the GPT4All library to execute LLaMa2, LLaMa3, Orca, and Falcon. We used the default temperature value. Based on the documentation, the temperature was set to 0.7. For GPT3.5 and 4, the temperature setting was 0.7 (currently, the default value for the API for GPT4o and newer is set to 1; \citealp{openai_openai_nodate}).

We used two prompts that were sent to each of the models. The first one is a simple ask for a term explanation in 10 sentences. The exact prompt is the following: \textit{Please describe in 10 sentences as plain text what <term> is}. The second prompt is a request for a text similar to the Wikipedia page. The exact prompt is the following: \textit{Please describe as it would be the Wikipedia page in 10 sentences what <term> is}. The reason for having two prompts is that the term explanation can be potentially easier to be recognized when compared with a model-generated text. That is why the Wikipedia page-like response is compared. 

Table~\ref{tab:dataset_stats} shows basic statistics of the dataset. In particular, the GPT models and LLaMa~3 kept very close to the 10-sentence limit, while the other models tended to produce shorter sentences and paragraphs.

\begin{table*}[]
\begin{tabular}{lccccccc}
\toprule
    & \textbf{wiki} & \textbf{GPT-3.5} & \textbf{GPT-4} & \textbf{LLaMa~2} & \textbf{LLaMa~3} & \textbf{Orca} & \textbf{Falcon} \\ 
\midrule
\textbf{\#tokens}                      & 243 ± 56  & 223 ± 35    & 218 ± 27  & 152 ± 17    & 249 ± 35    & 144 ± 27  & 152 ± 39    \\
\textbf{fraction of punctuation [\%]} & 13.5 ± 3.3 & 11 ± 2.3    & 12.2 ± 2.3 & 11.3 ± 2.9   & 11.8 ± 2.3   & 12.3 ± 3.4 & 10.7 ± 3.1   \\
\textbf{\#tokens / sentence}           & 24.0 ± 5.5    & 23.1 ± 2.8      & 22.3 ± 2.6    & 22.0 ± 3.5      & 24.8 ± 3.1      & 22.3 ± 3.9    & 22.3 ± 3.5      \\
\textbf{\#sentences}                   & 10.19 ± 0.94  & 9.7 ± 1.2       & 9.78 ± 0.79   & 7.0 ± 1.3       & 10.04 ± 0.82    & 6.6 ± 1.7     & 7.0 ± 2.0       \\
\textbf{max. \#sentences}              & 18            & 18              & 14            & 16              & 17              & 16            & 17              \\ 
\bottomrule
\end{tabular}
    \caption{Basic LLM dataset statistics: total number of tokens (including punctuation), fraction of punctuation tokens in the total token count, mean number of tokens in a sentence, number of sentences, and the maximal number of sentences. The numbers are averages and standard deviations across all documents.}
    \label{tab:dataset_stats}
\end{table*}

\subsection{Benchmarks}
\label{sec:Benchmarks}

As external benchmark for machine-generated text detection we have used AuTexTification~\citealp{sarvazyan_overview_2023}.
AuTexTification contains two shared tasks in two languages (English and Spanish): (i) MGT detection (a binary classification of texts written by human and a language model) and (ii) MGT attribution (classification of six models). Importantly, the first task uses a balanced multi-domain (tweets, how-to articles, legal documents, reviews and news) and multi-model (BLOOM: BLOOM-1B7, BLOOM-3B, BLOOM-7B1, and GPT-3: babbage, curie, and text-davinci-003) corpus, where only the first three domains appear in the training data and the last two in the test data.



\begin{table*}[h!]
\renewcommand\cellalign{tl}
\renewcommand\theadalign{tl}

\begin{tabularx}{\textwidth}{@{}>{\raggedright\arraybackslash}p{3cm} c c >{\raggedright\arraybackslash}p{3.5cm} >{\raggedright\arraybackslash}p{2cm} >{\raggedright\arraybackslash}X@{}}
\toprule
\textbf{Benchmark} & \textbf{Human} & \textbf{LLMs} & \textbf{LLMs Type} & \textbf{Language} & \textbf{Domain} \\
\midrule
AuTexTification 1 & $\sim$28k & $\sim$28k & \makecell[l]{BLOOM-1B1, BLOOM-3B,\\ BLOOM-7B1, Babbage,\\ Curie, text-davinci-003} & English & \makecell[l]{tweets, how-to, news,\\ legal, reviews} \\
\bottomrule
\end{tabularx}

\caption{Benchmark used.}
\label{tab:benchmarks}
\end{table*}

\subsection{Stylometry}
\label{sec:Stylometry}

We use two stylometry libraries: StyloMetrix \citep{okulska2023stylometrix} and CLARIN-PL's stylometric pipeline \citep{ochab2024}. 

\subsubsection{StyloMetrix}

StyloMetrix is an open-source stylometric text analysis library. Covers various grammatical, syntactic, and lexical aspects. StyloMetrix allows allowing feature engineering and interpretability. Stylometry involves the analysis of linguistic features to characterize the style of texts. Previous tools like ‘stylo’ package in R \citep{eder2016RJStylometry} provide quantitative text analysis but lack certain metrics and usability features that StyloMetrix offers. It is based on the spaCy model for English and generates normalized vectors for input texts, allowing comparison across texts of different lengths and genres. Vectors are designed to be interpretable at different levels. Metrics that are available for the English language:
\begin{itemize}
    \item Detailed Grammatical Forms: Tenses, modal verbs, etc.
    \item General Grammar Forms: Consolidation of principal grammatical rules.
    \item Detailed Lexical Forms: Types of pronouns, hurtful words, punctuation, etc.
    \item Parts of Speech: General frequency calculation.
    \item Social Media: Sentiment analysis, lexical intensifiers, masked words, etc.
    \item Syntactic Forms: Questions, sentences, figures of speech, etc.
    \item General Text Statistics: Type-token ratio, text cohesion, etc.
\end{itemize}
The version of the library used in this paper provides 195 stylometry features. It also supports model explainability and is available in multiple languages, making it a valuable tool for linguistic analysis and machine learning applications.

\subsubsection{CLARIN-PL's stylometric pipeline}

We used a modular Python pipeline for interpretable stylometric analysis developed for CLARIN-PL\footnote{\url{https://gitlab.clarin-pl.eu/stylometry/cl_explainable_stylo}}\citep{ochab2024}. The pipeline connects text preprocessing and linguistic feature extraction with various NLP tools, classifiers, an explainability module, and visualization.
At present, we use spaCy~\citep{ines_montani_2023_10009823} model ‘en\_core\_web\_lg’ for preprocessing steps (including tokenisation, named entity recognition, dependency parsing, part-of-speech and morphology annotation), Light Gradient-Boosting Machine (LGBM)~\citep{ke2017lightgbm} as the state-of-the-art boosted trees classifier, Shapley Additive Explanations (SHAP)~\citep{lundberg2020local2global} for computing explanations, and Scikit-learn~\citep{scikit-learn} for feature counting and cross-validation.
The visualisation functions, showing general and detailed explanations of what linguistic features make texts differ, utilise spaCy and SHAP.

As in previous works \citep{argasinski2024,ochab2024}, we decided to use (i) tree models, which are easily interpretable and for which the explanations can be computed fast, (ii) feature engineering approach, where the features are rooted in linguistic knowledge but can be generated programmatically.
Specifically, the features passed to the classifier were the normalised frequencies of:
\begin{itemize}
    \item lemmas (from uni- to trigrams), excluding named entities,
    \item part-of-speech tags (from uni- to trigrams), excluding named entities and punctuation,
    \item dependency-based bigrams,
    \item morphological annotations (unigrams) excluding punctuation,
\end{itemize}
No culling (i.e., ignoring tokens with document frequency strictly higher or lower than the given threshold) was performed.
We specifically excluded punctuation marks after initial experiments, as the features containing them tended to express some of the Wikipedia preprocessing artefacts.
Such features can also be expressive of some artefacts in LLM processing, such as the `SPACE' token (a redundant whitespace character, e.g., at the beginning of a paragraph or a second one between words), as in the \hyperref[sec:Results]{Results}.
The whitespace token is used in the multiclass classification, but in the binary classification, we remove all 83 features containing it.

\subsection{Classification}
\label{sec:Classification}

The first method chosen is a simple decision tree classifier from the popular Python \texttt{sklearn}\footnote{\url{https://scikit-learn.org}} library. It was used with the default parameters such as the Gini impurity method, the minimum samples in the split set to 2, and the split strategy set to \textit{best}. The test and train sets were used in a split of 70\% to 30\% with a 10-fold cross-validation (CV).

The LGBM classifier was used with the following settings: DART boosting, maximal depth of the tree model ("max\_depth" = 5), maximal number of leaves per tree ("num\_leaves" = 5), default number of boosting iterations, increased "learning\_rate" = 0.5, enabled bagging (randomly selecting part of data without resampling with "bagging\_freq" = 3 and "bagging\_fraction" = 0.8), and number of classes in the multiclass scenario ("num\_class" = 7). Further hyperparameter optimisation is possible, but was not performed in this study.

We used the group cross-validation scheme by using 10-fold CV for test error estimation.
Group CV makes sure that a given topic of the summary never appears both in the train and test set.
The reported scores are averages over the CV loop.
Training and test set sizes in each fold were $4390$ and $488$ samples for binary classification and, respectively, $15365$ and $1708$ for multiclass classification.

For the binary classification scenario, we provide accuracy, since all the datasets are exactly balanced.
For the multiclass scenario, we provide the Matthews correlation coefficient (MCC) as the performance metric.
\begin{table*}[ht!]
    \centering
    \begin{tabular}{cccccccc}
 \toprule
 & \textbf{wiki} & \textbf{GPT-3.5} & \textbf{GPT-4} & \textbf{LLaMa~2} & \textbf{LLaMa~3} & \textbf{Orca} & \textbf{Falcon}\\
 \midrule
 \multicolumn{7}{c}{Prompt \#1}\\
 \midrule
 \textbf{wiki} & 1.0 &  0.8170 & 0.8693 & 0.9596 & 0.8324 & 0.9605 & 0.9286\\
 \textbf{GPT-3.5}  & & 1.0 & 0.7154 & 0.9263 & 0.6869 & 0.9273 & 0.8804\\
 \textbf{GPT-4} & & & 1.0 & 0.7740 & 0.5754 & 0.8124 & 0.7658\\
 \textbf{LLaMa~2} & & & & 1.0 & 0.8323 & 0.5693 & 0.6922\\
 \textbf{LLaMa~3} & & & & & 1.0 & 0.8525 & 0.8081\\
 \textbf{Orca} & & & & & & 1.0 & 0.6082\\
 \textbf{Falcon} & & & & & & & 1.0\\
 \midrule
 \multicolumn{7}{c}{Prompt \#2}\\
 \midrule
 \textbf{wiki} & 1.0 & 0.8230 & 0.8419 & 0.9451 & 0.7991 & 0.9475 & 0.9030\\
 \textbf{GPT-3.5} & & 1.0 & 0.6428 & 0.8884 & 0.6291 & 0.8905 & 0.8271\\
 \textbf{GPT-4} & & & 1.0 & 0.8380 & 0.5688 & 0.8501 & 0.8008\\
 \textbf{LLaMa~2} & & & & 1.0 & 0.8657 & 0.5256 & 0.6809\\
 \textbf{LLaMa~3} & & & & & 1.0 & 0.8778 & 0.8160\\
 \textbf{Orca} & & & & & & 1.0 & 0.6701\\
 \textbf{Falcon} & & & & & & & 1.0\\
 \bottomrule
    \end{tabular}
    \caption{Accuracy of binary text classification with decision trees. Each table entry corresponds to a task, where class 1 and 2 are column and row model labels, respectively. Texts generated by different prompts are analysed separately.}
    \label{tab:wikillm}
\end{table*}

\section{Results}
\label{sec:Results}

We have performed the classification on the same dataset using two different classifiers and two different stylometric libraries. For the sake of comparison, we also included the recognition of summarization methods with LLMs.

\subsection{Binary classification with decision trees}

The decision trees performed worse compared to LGBM. This was the first experiment to test if the models can be recognized between each other and the Wikipedia text. The results for two prompts explained in the previous section are given in Table \ref{tab:wikillm}.

Decision trees are known to be used to measure feature importance. In our first experiment the most significant stylometric features are as follows:
\begin{itemize}
    \item L\_ADJ\_COMPARATIVE -- adjectives in comparative degree,
    \item L\_FUNC\_T -- function words types,
    \item FOS\_FRONTING -- fronting,
    \item L\_TYPE\_TOKEN\_RATIO\_LEMMAS -- type-token ratio for words lemmas.
\end{itemize}
These four features were used for the binary classifications.
The worst results were achieved for the second prompt with the following pair of classes: Orca and LLaMa~2, LLaMa~3 and GPT-4, Falcon and LLaMa~2, and Falcon and Orca. In the first two cases the results were about 52\% and 56\% accordingly. We can conclude that in both cases the recognition is very limited or even fails. Majority of model binary recognitions are between 70\% and 85\%. The best results are for distinguishing LLaMa~2 from GPT-3.5, and Orca from GPT-3.5 for both prompts. The accuracy is about 92\%  for the first prompt, and about 89\% for the second prompt. What is worth attention are the results in recognition of models' generated text and the Wikipedia text where the lowest accuracy is about 73\%, but the majority is above 85\%, with best results achieved for Orca and LLaMa~2, 95\% and 96\% accordingly.

\subsection{Binary classification with LGBM}

\subsubsection{StyloMetrix features}

Table~\ref{tab:binary_gen_acc_SM} shows CV-averaged accuracy between all pairs of classes.
The LLM most often misclassified as the real Wikipedia are GPT-4 and LLaMa~3 (cf. Tables~\ref{tab:multiclass_gen_conf_SM}-\ref{tab:binary_gen_acc_SM}).
LLaMa~2 and Orca were the hardest to distinguish.
GPT models and LLaMa~3, as well as Orca and Falcon are also confused often.

\begin{table*}[ht!]
\centering
\begin{tabular}{cccccccc}
\toprule
 & \textbf{wiki} & \textbf{GPT-3.5} & \textbf{GPT-4} & \textbf{LLaMa~2} & \textbf{LLaMa~3} & \textbf{Orca} & \textbf{Falcon} \\ 
 \midrule
\multicolumn{8}{c}{\textbf{Stylometrix features}}\\
\midrule 
\textbf{wiki}   &    & 0.97    & 0.94  & 0.99    & 0.95    & 0.99  & 0.98    \\
\textbf{GPT-3.5} &    &      & 0.87  & 0.99    & 0.88    & 0.99  & 0.98    \\
\textbf{GPT-4}   &    &      &    & 0.99    & 0.85    & 0.99  & 0.98    \\
\textbf{LLaMa~2} &    &      &    &      & 0.99    & 0.77  & 0.90    \\
\textbf{LLaMa~3} &    &      &    &      &      & 0.99  & 0.98    \\
\textbf{Orca}   &    &      &    &      &      &    & 0.87    \\
\textbf{Falcon} &    &      &    &      &      &    &      \\ 
\midrule
\multicolumn{8}{c}{\textbf{Frequency-based features}}\\
\midrule
\textbf{wiki}   &  & 0.99    & 0.98  & 1.00    & 0.99    & 1.00  & 1.00    \\
\textbf{GPT-3.5} &  &    & 0.90  & 0.98    & 0.91    & 0.98  & 0.97    \\
\textbf{GPT-4}   &  &    &  & 0.99    & 0.93    & 0.99  & 0.98    \\
\textbf{LLaMa~2} &  &    &  &    & 0.99    & 0.79  & 0.84    \\
\textbf{LLaMa~3} &  &    &  &    &    & 1.00  & 0.99    \\
\textbf{Orca}   &  &    &  &    &    &  & 0.86    \\
\textbf{Falcon} &  &    &  &    &    &  &    \\
\bottomrule
\end{tabular}
    \caption{Accuracy of binary text classification with LGBM using StyloMetrix features. Each table entry corresponds to a task, where class 1 and 2 are column and row model labels, respectively. The results are averages over 10 CV folds.}
    \label{tab:binary_gen_acc_SM}
\end{table*}

\subsubsection{Frequency-based features} 

Table~\ref{tab:binary_gen_acc_SM} shows the accuracy between all pairs of classes.
LLMs are hardly confused with the real Wikipedia at all. 
As before, the most often confused pairs of models were GPT models and LLaMa~3, as well as the triplet LLaMa~2, Orca, and Falcon.


\subsection{Multiclass classification with LGBM}

The performance of LGBM classifier is reported in Table~\ref{tab:multiclass_gen}.
Visibly, it heavily depends on the number and selection of the features used.
The small variance of the results across CV folds indicates that the results are robust.

\begin{table}[h!]
\centering
    \begin{tabular}{ccc}
 \toprule
 & \textbf{StyloMetrix} & \textbf{Frequencies} \\
 \midrule
 \textbf{CV average} & 0.72 & 0.87\\ 
\textbf{CV min.} & 0.71 & 0.86\\
\textbf{CV max.} & 0.74 & 0.89\\
\textbf{dummy baseline} & 0.00 & 0.00 \\
\textbf{number of features} & 196 & 3000 \\
 \bottomrule
    \end{tabular}
    \caption{Multiclass generators performance [MCC].}
    \label{tab:multiclass_gen}
\end{table}


\subsubsection{StyloMetrix features}

Table~\ref{tab:multiclass_gen_conf_SM} shows the normalised confusion matrix.
Interestingly, the man-made Wikipedia texts are recognised better than any of the LLMs.
The largest confusion exists between LLaMa~2 and Orca models and between LLaMa~3 and the GPT models.
The LLM most often misclassified as the real Wikipedia is GPT-4.

\begin{table*}[ht!]
    \centering
    \begin{tabular}{cccccccc}
    \toprule
    & \textbf{wiki} & \textbf{GPT-3.5} & \textbf{GPT-4} & \textbf{LLaMa~2} & \textbf{LLaMa~3} & \textbf{Orca} & \textbf{Falcon} \\
    \midrule
 \multicolumn{8}{c}{\textbf{Stylometrix features}}\\
    \midrule 
    \textbf{wiki} & 0.90 & 0.011 & 0.040 & 0.0078  & 0.030   & 0.0062 & 0.0082  \\
    \textbf{GPT-3.5} & 0.017 & 0.78    & 0.089 & 0.0041  & 0.094   & 0.0090 & 0.0082  \\
    \textbf{GPT-4}   & 0.044 & 0.11    & 0.73  & 0.0082  & 0.10    & 0.0029 & 0.0090  \\
    \textbf{LLaMa~2} & 0.0082 & 0.0033  & 0.0057 & 0.72    & 0.0033  & 0.19  & 0.071   \\
    \textbf{LLaMa~3} & 0.044 & 0.097   & 0.11  & 0.0049  & 0.74    & 0.0016 & 0.0082  \\
    \textbf{Orca}   & 0.013 & 0.0049  & 0.0033 & 0.22    & 0.0037  & 0.67  & 0.085   \\
    \textbf{Falcon} & 0.011 & 0.011   & 0.0082 & 0.078   & 0.011   & 0.087 & 0.79    \\
    \midrule
\multicolumn{8}{c}{\textbf{Feature-based features}}\\
    \textbf{wiki}   & 0.98  & 0.0012  & 0.011 & 0.0     & 0.0033  & 0.0016 & 0.0     \\
    \textbf{GPT-3.5} & 0.0037 & 0.83    & 0.078 & 0.00041 & 0.063   & 0.016 & 0.0057  \\
    \textbf{GPT-4}   & 0.015 & 0.069   & 0.85  & 0.0025  & 0.041   & 0.011 & 0.0074  \\
    \textbf{LLaMa~2} & 0.0   & 0.0     & 0.0   & 0.96    & 0.0     & 0.011 & 0.031   \\
    \textbf{LLaMa~3} & 0.015 & 0.065   & 0.048 & 0.00082 & 0.85    & 0.0029 & 0.015   \\
    \textbf{Orca}   & 0.00082       & 0.0041  & 0.0049 & 0.014   & 0.0     & 0.88  & 0.097   \\
    \textbf{Falcon} & 0.0012 & 0.0057  & 0.0033 & 0.0094  & 0.0029  & 0.11  & 0.87    \\
    \bottomrule
\end{tabular}
    \caption{Confusion matrix in the multiclass classification scenario for LGBM using StyloMetrix and frequency-based features.}
    \label{tab:multiclass_gen_conf_SM}
\end{table*}

\subsubsection{Frequency-based features}

Table~\ref{tab:multiclass_gen_conf_SM} shows the normalised confusion matrix.
Again, Wikipedia has the highest accuracy and the LLM most often misclassified as it is GPT-4.
The most often confused pairs of models are Falcon and Orca, GPT-3.5 and GPT-4, LLaMa~3 and GPT-3.5.

\subsection{Robustness testing}

For brevity, we provide robustness testing only for the LGBM model with frequency-based features,
and only for the binary detection of human- and machine-generated texts.
Since the test sets contains only one class (machine-generated texts), we provide the value of recall of that class and the validation recall for comparison, see Table~\ref{tab:robust}.

\subsubsection{Testing on unseen models}
The test assumes that in training the model can only access data on man-made texts and on five out of six LLMs. The features are chosen and fixed at this stage and the training recall is computed.
Testing is performed on the single LLM previously unseen by the model.
The cross-validation here regards the training only, i.e., for each unseen LLM there were 10 classifiers trained on subsets of the training set (and evaluated on the validation set as shown in Table~\ref{tab:robust}), while the test set remained the same.
The standard deviations of recall are computed over these 10 folds.

The largest drop in performance can be seen for GPT-4 and LLaMa~3 models.

\begin{table*}[ht!]
\centering
\begin{tabular}{rcccc}
\toprule
\textbf{Recall {[}\%{]}} & \textbf{Validation} & \textbf{Test} & \textbf{DIPPER} & \textbf{Parrot} \\
\midrule
\textbf{GPT-3.5}         & 99.11 ± .36    & 99.6 ± .17    & 99.95±.062      & 99.971 ± 0.044  \\
\textbf{GPT-4}           & 99.49 ± .24    & 88.2 ± 1.2    & 99.95±.045      & 98.81 ± 0.14    \\
\textbf{LLaMa~2}         & 99.17 ± .39    & 99.61 ± .11   & 99.922±.065     & 99.992 ± 0.017  \\
\textbf{LLaMa~3}         & 99.24 ± .24    & 94.13 ± .72   & 99.9±.064       & 99.736 ± 0.098  \\
\textbf{Orca}            & 99.18 ± .32    & 99.79 ± .16   & 99.87±.17       & 99.996 ± 0.013  \\
\textbf{Falcon}          & 99.14 ± .30    & 99.691 ± .056 & 99.81±.11       & 99.955 ± 0.03   \\
\bottomrule
\end{tabular}
    \caption{Recall on (i) unseen LLMs and on texts paraphrased with (ii) DIPPER and (iii) Parrot. (i) Unseen LLM test set contained only the given model, while training set contained the other models and the human Wiki summaries. (ii)-(iii) Paraphrased test set contained only the paraphrases, while the training set contained all the models and human texts. The values are mean and standard deviation over CV folds.}
    \label{tab:robust}
\end{table*}

\subsubsection{Testing on paraphrased texts}
Following \cite{sadasivan_can_2025} we performed a paraphrase attack using DIPPER \cite{krishna_paraphrasing_2023}, a 11B paraphrasing model, and Parrot~\cite{damodaran_parrot_2021}, a T5-based paraphrase model.
Reportedly, in a small sample, DIPPER had shown in human evaluation that the content was satisfactorily preserved in about 70\% of the samples and grammar quality was satisfactory in 88\%.
We used no recursive paraphrasing. 

The test assumes that, in training, the model can access all unparaphrased data (human and all six LLM-generated texts). The features are chosen and fixed at this stage.
Testing is performed on all paraphrases.
As above, the cross-validation here regards the training only.
The results are shown in Table~\ref{tab:robust}.

In general, paraphrasing resulted in a higher detection rate than for unparaphrased texts. The only exception is Parrot paraphrasing GPT-4, where a less than 1\% drop in recall occurred.

\subsubsection{Testing on cross-domain benchmark}

The results obtained on the AuTexTification task are shown in Table \ref{tab:benchmark_res}.
Our classifiers used only the training data provided in the task description according to the task constraints. 

\begin{table}[h!]
\centering
    \begin{tabular}{rc}
 \toprule
 \textbf{Classifier} & \textbf{F1} \\
 \midrule
 Top & 0.81 \\
 LR &	0.66 \\
 \cite{mikros2023transformers} & 0.61 \\
 StyloMetrix & 0.48 \\
 Frequencies & 0.54 \\
 \bottomrule
    \end{tabular}
    \caption{AuTexTification benchmark results. Macro-F1 score. For comparison submissions to \cite{sarvazyan_overview_2023} are presented: the top-ranked, the logistic regression (LR) baseline and results by \cite{mikros2023transformers} (an ensemble of stylometric features and transformers).}
    \label{tab:benchmark_res}
\end{table}

\subsection{Explainability}
This section reports the results of SHAP explanations. 
In the binary classification, we provide only a single example to show the effectiveness of the explanations. For that purpose the GPT-4 model was chosen as the hardest one to detect. In this case, the positive or negative direction of SHAP values points toward one or the other class.
In the multiclass classification, the obtained explanations take into account all the LLMs. In that case, the absolute values of SHAP show which features explain which model to what extent. Depending on the model's idiosyncrasies, a feature can explain several models well, but others poorly. Moreover, some models may be explained by few very strong features (i.e., with large SHAP values), while others may need numerous features contributing only small fractions to the explanation, as visible in Figure \ref{fig:shaps_multiclass}.

For each classification scenario, SHAPs were collected and averaged across all CV folds.

\subsubsection{Binary classification}

\begin{figure*}[ht!]
    \centering
    \subfigure[]{
    \includegraphics[width=0.48\textwidth]{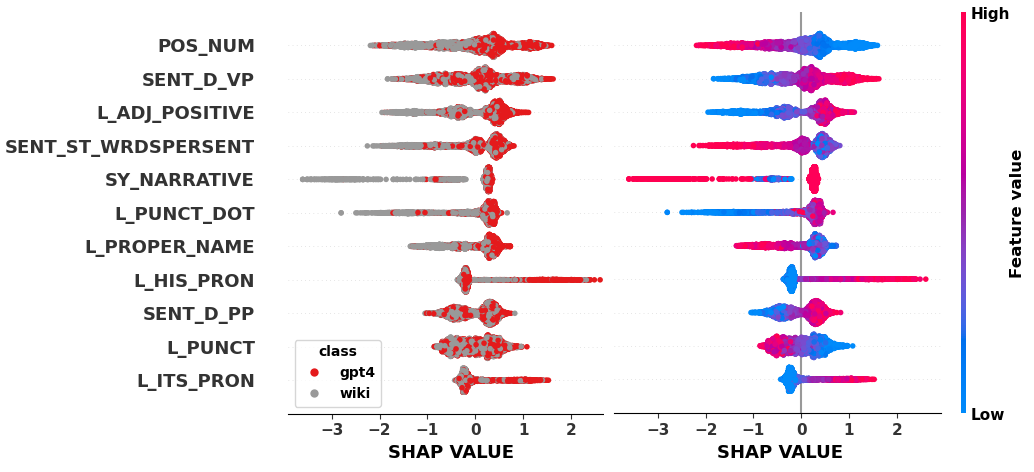}
    }
    \subfigure[]{
    \includegraphics[width=0.48\textwidth]{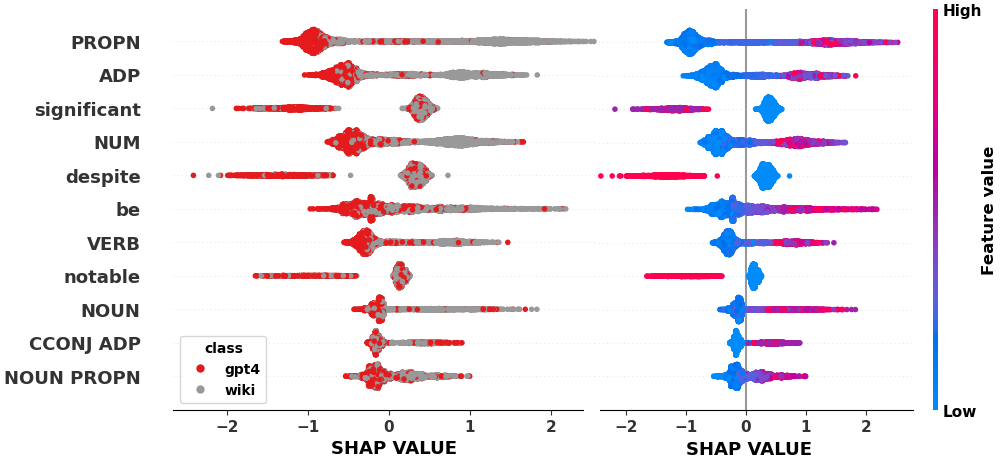}    
    }
    \caption{Explanations for binary classification between the Wikipedia and GPT-4 using (a) StyloMetrix and (b) frequency-based features. Only the first 10 most important features are shown. Each point is a 10-sentence sample describing a given term coloured by: (\textbf{Left}) the sample's class, and (\textbf{Right}) its feature's intensity. The left plots indicate whether positive or negative SHAPs point toward GPT or the real Wikipedia.}
    \label{fig:shaps_binary_SM}
\end{figure*}

Here we present only the example of classifying the Wikipedia and GPT-4, as shown in Figures~\ref{fig:shaps_binary_SM}(a) and (b), respectively, for StyloMetrix and frequency features.
Analogous analyses can be repeated for the other pairs of classes.
Let us recall, that punctuation (including the SPACE token) was excluded from the frequency features.
Like above in the multiclass scenario, one notices features representing proper names (L\_PROPER\_NAME, PROPN), dates and other numerals (POS\_NUM, NUM), etc.
GPT-4 strikingly tends to abuse words like `significant', `notable' or `despite'.
Its usage of grammatical features (i.e., POS n-grams), however, tends to be strongly frequency-standardised, visible as the red bulks of the distributions in contrast to the long grey outlying distributions for the Wikipedia.

\subsubsection{Multiclass classification}
\begin{figure}[ht!]
    \centering
    \includegraphics[width=0.5\textwidth]{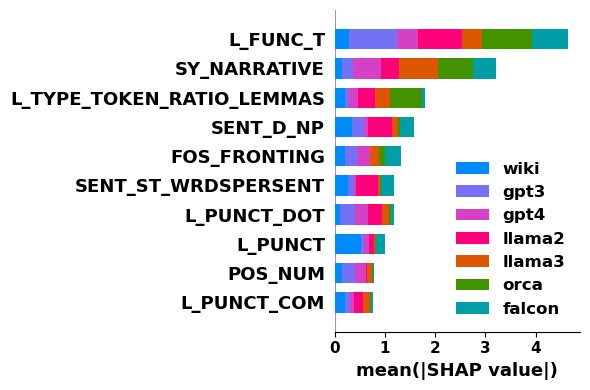}
    \vfill
    \includegraphics[width=0.5\textwidth]{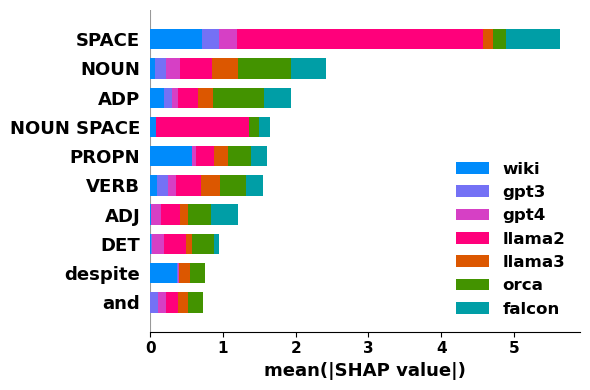}
    \caption{General explanations for multiclass classification. The first 10 most important features according to the absolute values of SHAP are shown. SHAP values were averaged over CV folds. Colours indicate the importance of a feature for recognising a particular class.}
    \label{fig:shaps_multiclass}
\end{figure}

In Fig.~\ref{fig:shaps_multiclass} the ten most important StyloMetrix and frequency features are shown.

The StyloMetrix features include (in the order of importance):
number of function word types,
number of words in narrative sentences,
the type-token ratio for word lemmas,
statistics between noun phrases,
fronting,
difference between the number of words and the number of sentences,
punctuation -- dots,
punctuation,
punctuation -- commas, and
numerals; see \citep{okulska2023stylometrix} for feature descriptions.
The frequency features include single part-of-speech tags such as:
whitespace, nouns, adpositions, proper nouns, verbs, adjectives, and determiners; POS bigrams such as: noun followed by a whitespace; and single lemmas such as: ‘despite’, ‘and’.

\begin{figure*}[ht!]
    \centering
    \begin{minipage}{.48\linewidth}
        \makebox[\linewidth]{\textbf{Stylometrix}}
    \subfigure[]{
    \includegraphics[width=\textwidth]{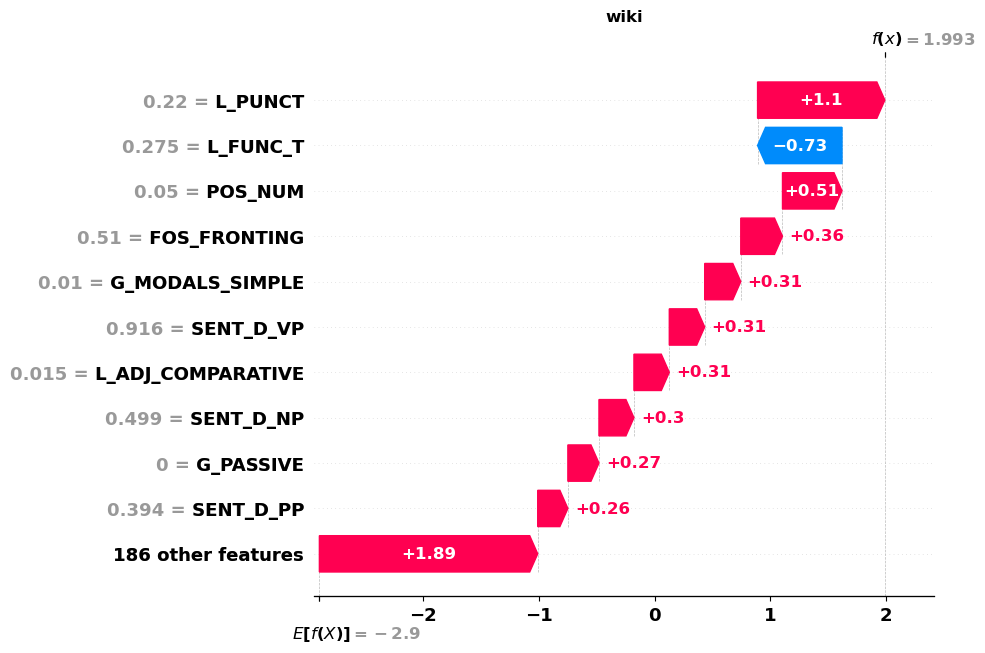}
    }
    \hfill
    \subfigure[]{
    \includegraphics[width=\textwidth]{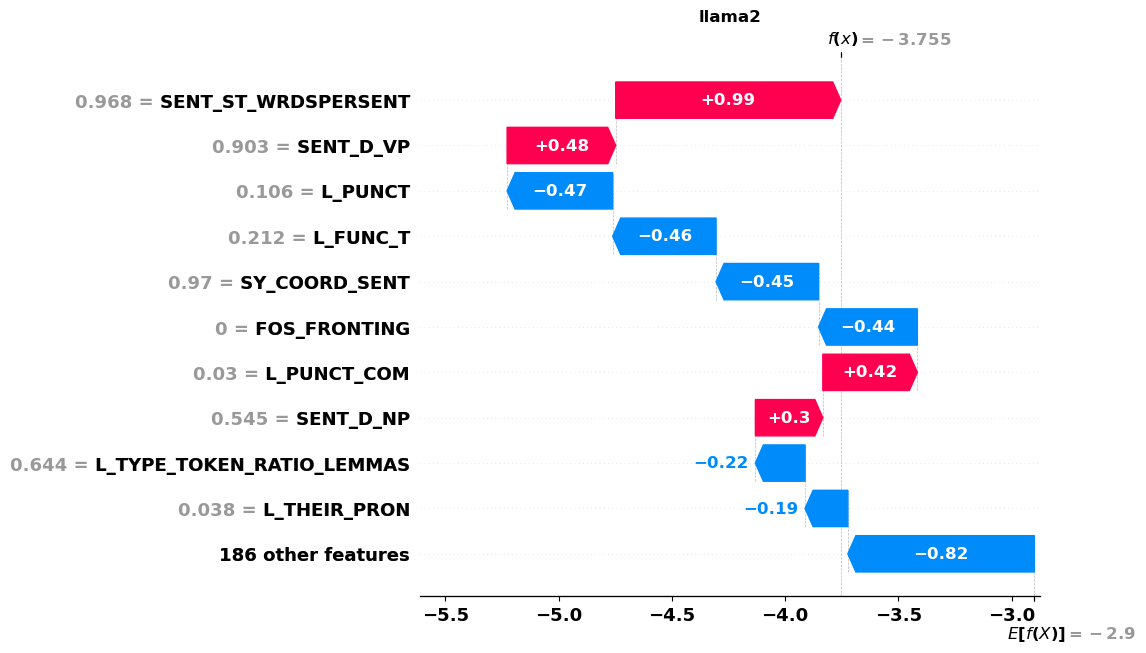}
    }
    \subfigure[]{
    \includegraphics[width=\textwidth]{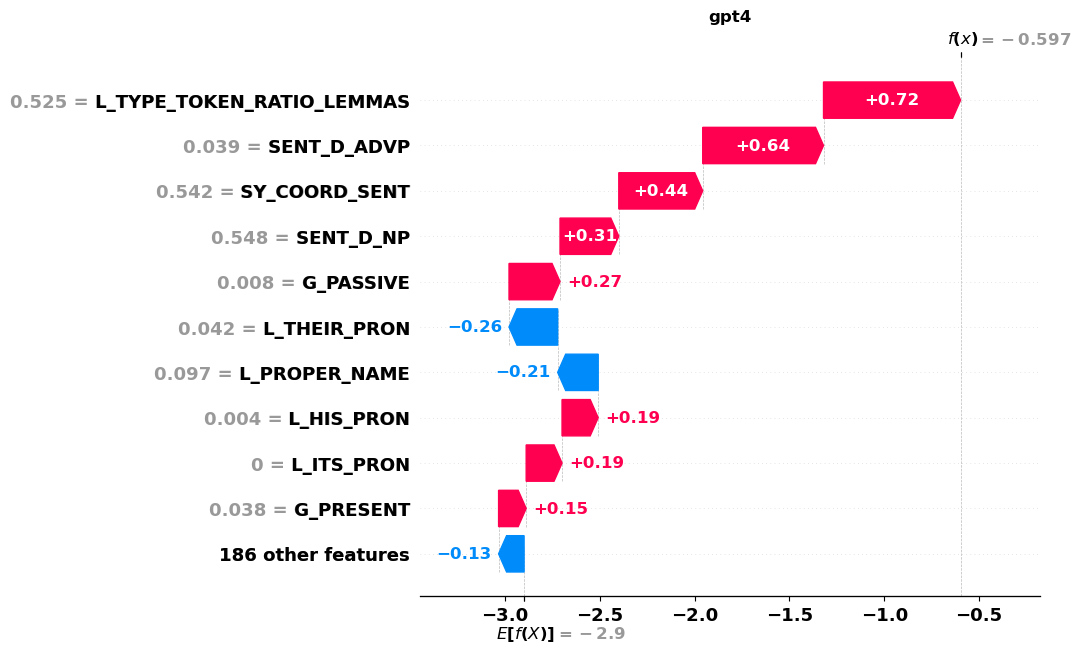}

    }    
    \end{minipage}
    \begin{minipage}{.48\linewidth}
    \makebox[\linewidth]{\textbf{Frequency-based}}
    \subfigure[]{
        \includegraphics[width=\textwidth]{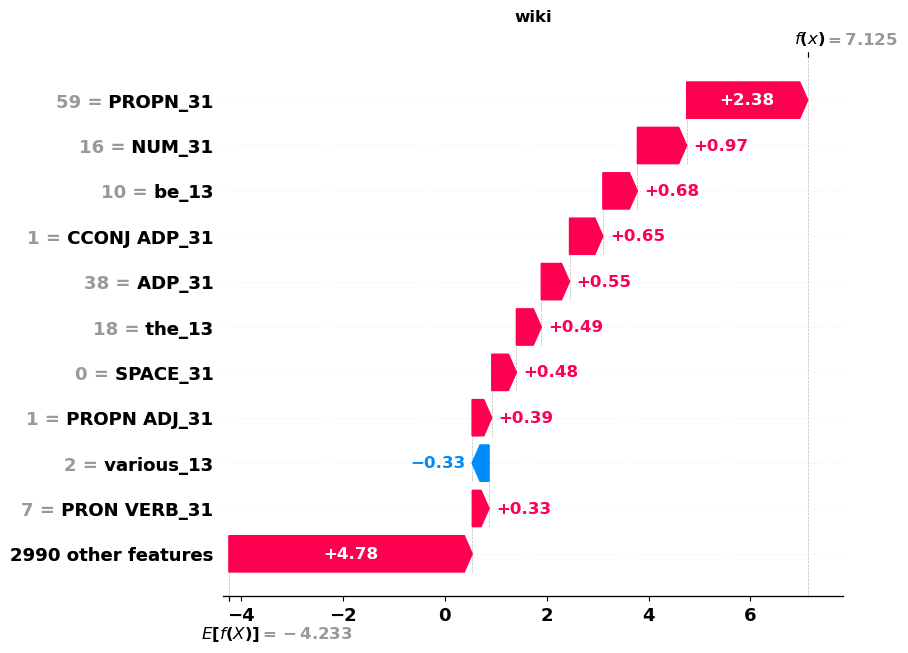}
    }
    \subfigure[]{
        \includegraphics[width=\textwidth]{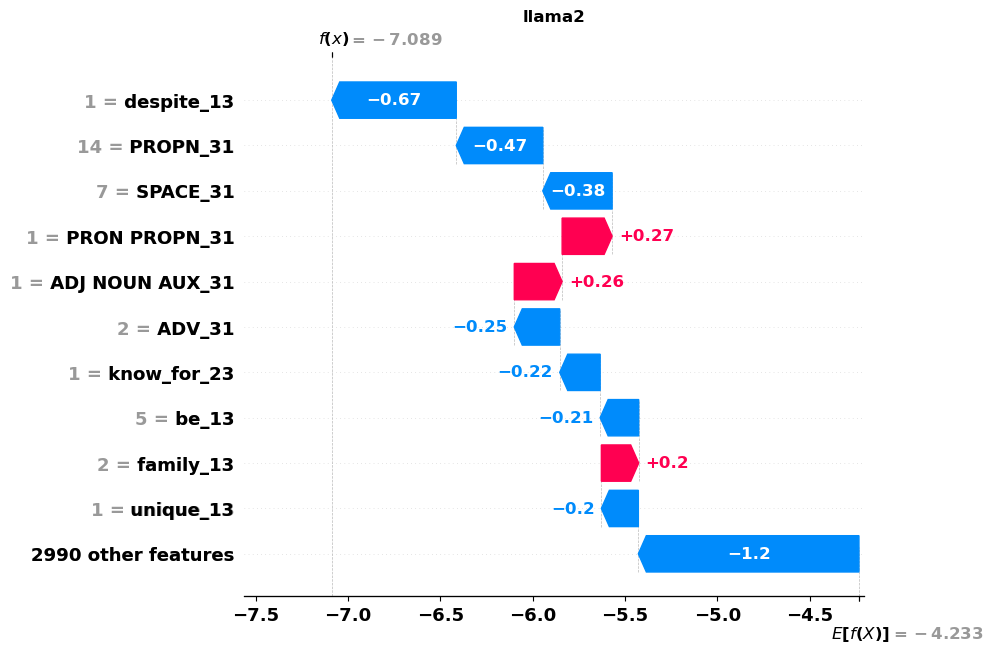}
    }
    \subfigure[]{
      \includegraphics[width=\textwidth]{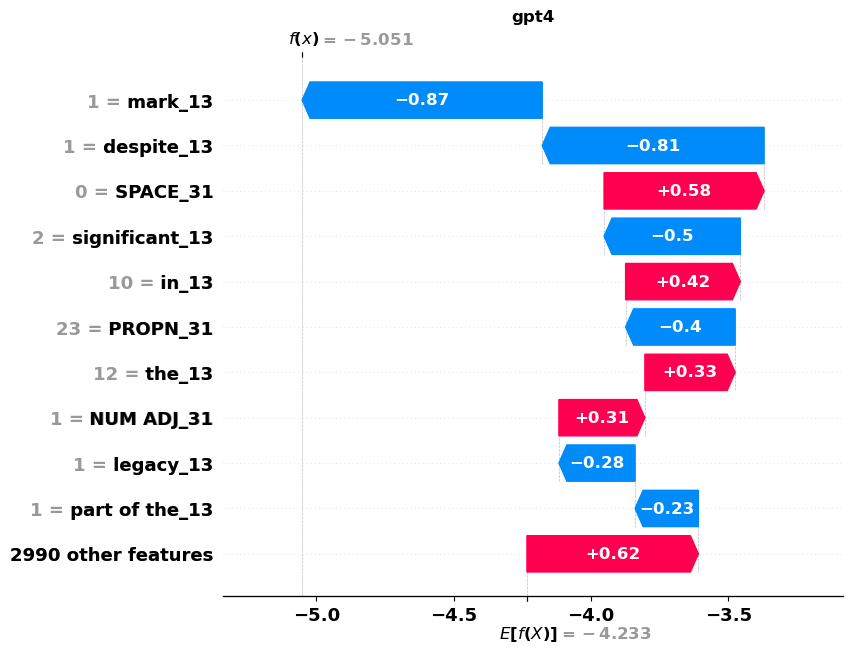}
    }
    \end{minipage}
    \caption{Local explanations of 10 most important StyloMetrix (a-c) and frequency features (d-f) in multiclass classification for text samples describing the term `The Swarbriggs'. Only selected models are shown. For this term, the Wikipedia was classified correctly,  GPT-4 was misclassified as the Wikipedia, and LLaMa~2 was misclassified as Orca. Grey numbers to the left indicate feature values in this particular text sample. The positive/negative SHAP values do not point strictly to any particular class (in the multiclass scenario) but they tend to be higher for Wikipedia and GPT models and lower for worse models.}
    \label{fig:shaps_multiclass_example_SM}
\end{figure*}

\begin{figure}[ht!]
    \centering
    \includegraphics[width=0.5\textwidth]{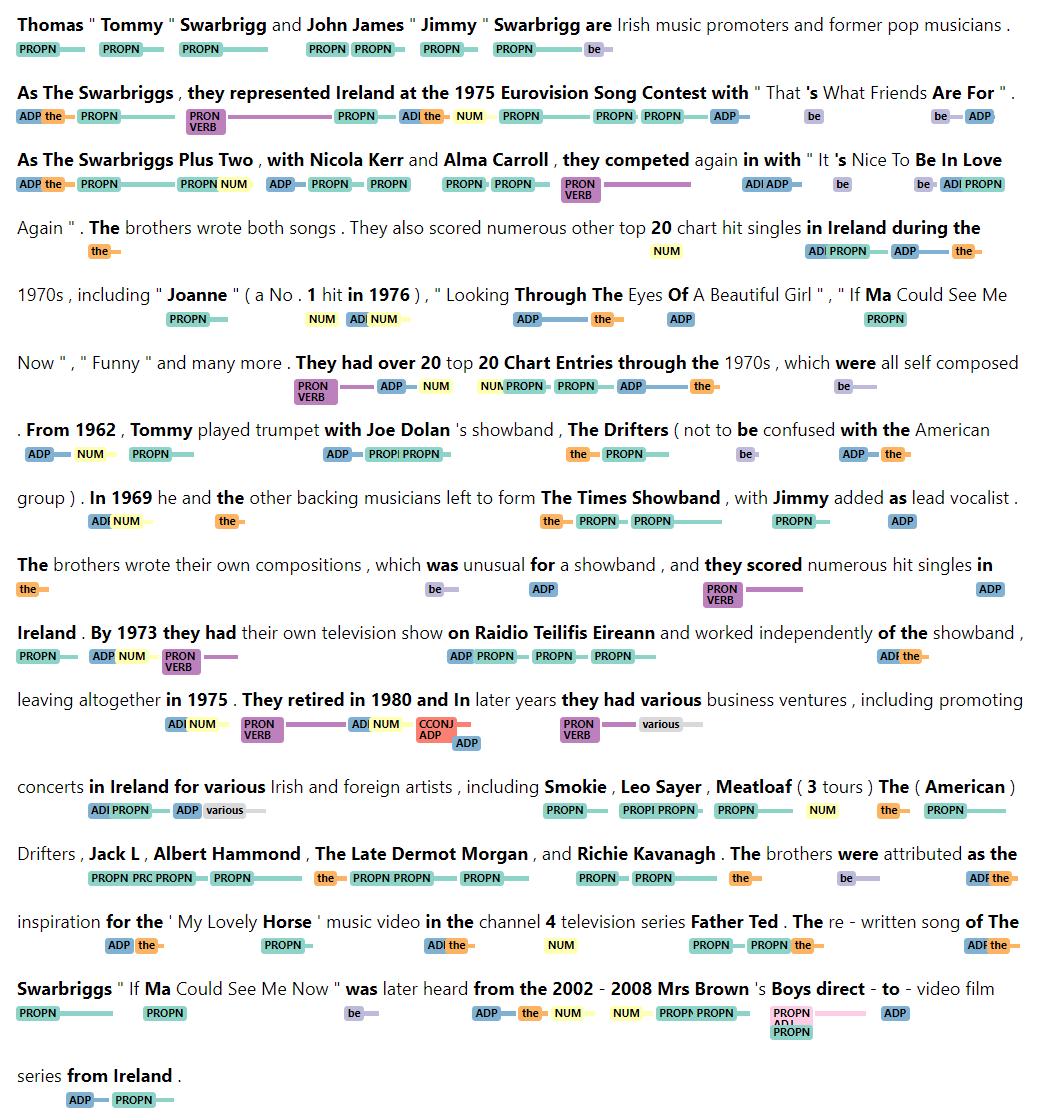}
    \caption{Text sample from the the Wikipedia with highlighted text spans corresponding to important frequency features from Fig.~\ref{fig:shaps_multiclass_example_SM}. Note that the lack of features (like SPACE) cannot be highlighted but is important to the classifier.}
    \label{fig:shaps_multiclass_textWIKI_FF}
\end{figure}
\begin{figure}[h!]
    \centering
    \includegraphics[width=0.48\textwidth]{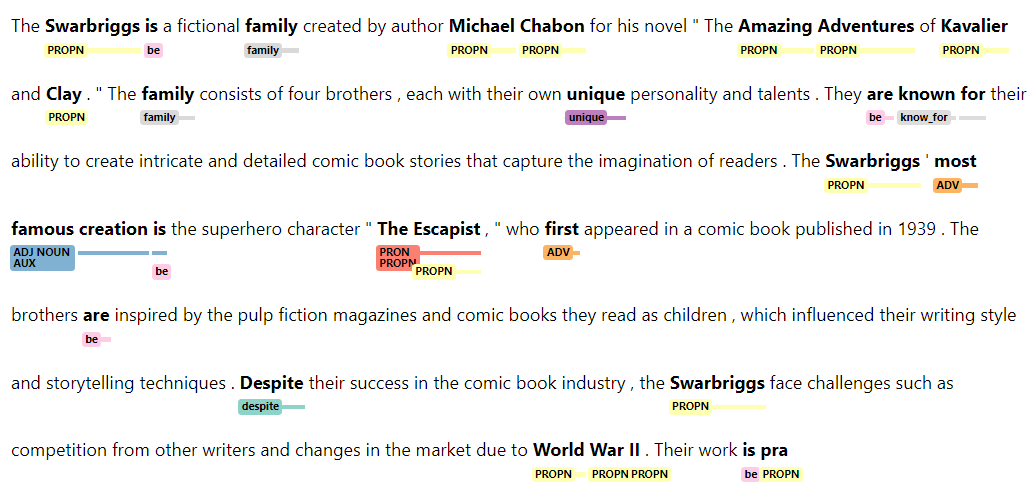}
    \caption{Text sample from LLaMa~2 with highlighted important frequency features.}
    \label{fig:shaps_multiclass_textLLAMA2_FF}
\end{figure}

\begin{figure}[h!]
    \centering
    \includegraphics[width=0.48\textwidth]{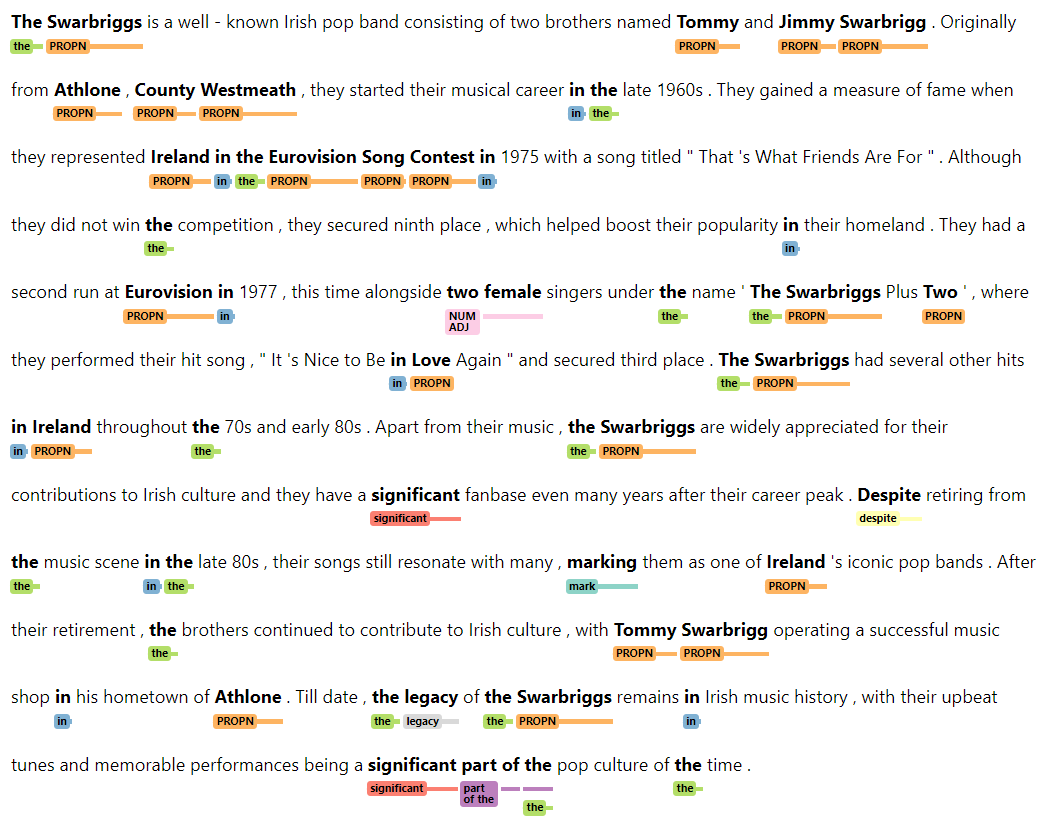}
    \caption{Text sample from GPT-4 with highlighted important frequency features.}
    \label{fig:shaps_multiclass_textGPT4_FF}
\end{figure}

The feature explanations are not model-agnostic. 
Notice the dates in the Wikipedia sample (POS\_NUM), lower number of punctuation marks for LLaMa~2 than for Wikipedia (see numbers next to L\_PUNCT in Fig.~\ref{fig:shaps_multiclass_example_SM}a), SENT\_D\_NP having similar values in all three cases. Also, looking at Fig.~\ref{fig:shaps_multiclass_example_SM}b, one notices a significantly larger number of proper nouns and dates in Wikipedia (PROPN -- also in bigrams -- and NUM), redundant spaces in LLaMa~2 (SPACE), and other singular features.
One can notice that the explanations for the GPT models are more distributed (no single feature with a huge SHAP value) than the other models.

It is worth recalling that models trained on different data subsets (CV folds) contribute to the SHAP values in Fig.~\ref{fig:shaps_multiclass}, while the SHAP values presented in Fig.~\ref{fig:shaps_multiclass_example_SM} correspond to a single classifier whose test set contained the selected texts.
In addition, the SHAP values in Fig.~\ref{fig:shaps_multiclass_example_SM} are averaged over classes, but explanations for each class can be obtained separately. 

Text samples (corresponding to the term `The Swarbriggs') are shown in Fig.~\ref{fig:shaps_multiclass_textWIKI_FF}-\ref{fig:shaps_multiclass_textGPT4_FF}, where also the frequency features most important to the classifier have been marked.

\subsection{Summarization methods comparison}

The text summarization methods are used only for comparison reasons as to what popular methods perform in a binary classification against language models. Similarly to LGBM experiments, this was also performed on the first prompt, because there were no significant differences between both prompts in the decision tree classification. The classification of summarization methods was also performed using the decision tree method. The results are given in Table \ref{tab:summarizers}.

\begin{table*}[ht!]
     \centering
     \begin{tabular}{p{1.5cm}p{2cm}cccc}
 \toprule
& \textbf{Wikipedia sum} & \textbf{Sumy} & \textbf{T5}  & \textbf{BART}  & \textbf{Gensim} \\
  \midrule
\textbf{Sumy} & 0.7540 & 1.0 & & \\
\textbf{T5} & 0.864 & 0.9221 & 1.0 & \\
\textbf{BART} & 0.9664 & 0.9735 & 0.9381 & 1.0 \\
\textbf{Gensim} & 0.9611 & 0.9753 & 0.948 & 0.7954 & 1.0\\
\textbf{GPT-3.5} & 0.7540 & 0.8398 & 0.8889 & 0.9648 & 0.9634 \\
\textbf{GPT-4} & 0.7283 & 0.8071 & 0.8967 & 0.9501 & 0.9469 \\
\textbf{LLaMa~2} & 0.8924 & 0.9129 & 0.9034 & 0.8135 & 0.8986 \\
\textbf{LLaMa~3} & 0.6865 & 0.79 & 0.8757 & 0.9622 & 0.9509\\
\textbf{Orca} & 0.9046 & 0.9223 & 0.9107 & 0.7561 & 0.8717 \\
\textbf{Falcon} & 0.8362 & 0.8875 & 0.8353 & 0.7935 & 0.8769\\
  \midrule
     \end{tabular}
     \caption{Accuracies of summarization methods text generation recognition using decision trees. Average over 10 CV folds.}
     \label{tab:summarizers}
 \end{table*}
The worst-recognized model is GPT-4, as the comparison with the Wikipedia summary is only about 72\%. This indicates that this model can simulate the way human summarizes Wikipedia pages, but it is important to highlight that it was also the most complex model used in our experiment. The other questionable recognitions were obtained for Orca compared to BART summarizer and Sumy summarizer compared to Wikipedia, about 75\% both. The other results vary between 80\% and 92\%. The best results were achieved in a binary classification to recognize the text generated by one summarizing methods from another summarizing method or LLM, like T5 summarizer and Sumy summarizer -- with about 92\%, Sumy summarizer and Orca -- about 92\%, GPT-4 and BART summarizer -- about 95\%, LLaMa~3 and BART summarizer -- about 96\%, BART summarizer and Wikipedia -- about 96\%, and BART summarizer and Sumy summarizer -- about 97\%.
\section{Discussion}
\label{sec:Discussion}

Generally, the results show that in a well-defined text generation task LLMs can be easily distinguished from the man-made texts and from each other with a boosted tree classifier even with very few features (196 for StyloMetrix in English) and even for extremely short texts (10 sentences). More features, coming mostly from grammatical tagging, lead to even better  -- indeed, almost perfect -- results.

From multiclass explanations: it seems that well-performing models do not have single strongly recognisable features, but their style is more dispersed among many quantified features.
Moreover, the explanations are not general, but may vary depending on the model, hence, the multiclass training is indispensable.
These plots summarise all folds in the cross-validation loop, so the results are also stable in terms of different training/test splits.
Interestingly, simple features such as the number of punctuation marks matter.
The whitespaces found in LLaMa~2 were actually double spaces between tokens or a space at the beginning of the text.
The number of full stops appears as a distinguishing feature, possibly because the LLMs tend to stop generating the text in the middle of the sentence. This might also affect `the difference between the number of words and the number of sentences' (SENT\_ST\_WRDSPERSENT) as well as some other features.
Wikipedia descriptions tend to be more fact-packed (dates and proper nouns) than LLM-generated ones. The distributional plots from binary classification between Wikipedia and GPT-4, suggest that the LLM favours certain individual words and is more standardised than Wikipedia in terms of grammatical structures (represented by frequencies of part-of-speech n-grams) -- perhaps an expected outcome since the Wikipedia text samples were authored by many people.
These conclusions come from explanations collected in the cross-validation loop, so they are stable in terms of different training/test splits.

The summarisation methods achieve similar results in the decision tree experiment. We can conclude that we will achieve similar results in LGBM for the summarisation methods. It indicates that the summarisers do have a distinctive way of text summarisation that can be found using stylometry.
Succinctness of BART and artefacts in T5 explains their high recognisability. Sumy is the most successful due to its flexibility in choosing the summary length.

\subsection{Limitations}

The limitations of the present paper concern mainly the material of the analysis.
Firstly, the results and specific conclusions refer only to the chosen text type, i.e., introductions to Wikipedia articles, which are expected to conform to an encyclopaedic style: plain, factual and partly formulaic.
Some of the most distinctive features reflect that, and cannot be generalised to classifying other text types.
However, the analytic pipeline is generic, including the engineered features, which have been designed and used in the context of literary texts.
Whether the cross-domain classification with this type of model is robust is at this time debatable, taking into account our preliminary results in the AuTexTification task~\citealp{sarvazyan_overview_2023}, but also results of others that have tried utilising stylometric features \citealp{mikros2023transformers} with results below simple baselines.
One can frame the issue of domain dependence of the model in various ways: both training and testing in another domain or cross-domain detection (i.e., detection on unseen domains), and which part of the data is unseen (whether the human- or machine-generated texts or both). Depending on these choices, the attack scenario is more or less realistic. 

Secondly, the language of the text samples is limited to English only.
The precise lexical, grammatical and other complex features will differ for other languages.
Performance of stylometric tools has been known to depend heavily on language and specifically on language type (analytic, synthetic, etc.), see, e.g. \citep{eder_style-markers_2011, evert_understanding_2017}. However, the LLMs are also best developed in English~\citep{li_quantifying_2024} and hence we expect it to be the most challenging setting for classification.
The text processing pipeline we used strictly depends on the availability of NLP tools (like POS taggers, dependency parsers, NERs, etc.) for a given language. 
The frequency features at this moment depend on spaCy, which currently provides more or fewer tools for about 24 languages.
In the case of StyloMetrix features, even though they also depend on the models distributed by spaCy, they were custom-designed for Polish, English, German, Ukrainian and Russian only.

Thirdly, the collection of Wikipedia samples is multi-authorial in at least two ways: each article could have been written by a different author, but also a single article probably has been edited by several authors -- of various individual styles and linguistic competency. Reproducing this variety has not been explicitly stated in any of the prompts.

Besides the single domain constraint, the robustness testing of the stylometric detection methods and their explanations is limited in terms of the variability of LLM generation. One can envisage generating multiple text versions with: varying hand-crafted or machine-paraphrased prompts, persona-assigned prompts~\cite{przystalski_building_2025,liu_evaluating_2024,wang_towards_2024}, as well as the same prompt with varying LLM parameters.
In our Wikipedia-based dataset, however, we do not expect much variance by varying the prompts, due to the constraints of the encyclopaedic style. In this case, we consider varying and tuning the prompts a less realistic attack scenario.

One should also note, that the multiclass classification is performed on a closed set of classes. Although adding unseen models does not change the task in binary classification (human vs. machine), in the multiclass case the task would change to an open set problem.

While the binary classification task (human vs. machine detection) remains unchanged with the addition of new generative models, the multiclass setting fundamentally changes: the task becomes an open-set classification problem~\cite{geng_recent_2020}, where the classifier additionally has to recognize samples that belong to unknown or novel classes. This issue is out of scope of the present paper, however, having a good close-set classifer is helpful in the open-set problems~\cite{vaze_open-set_2022}.

The language and type of the human-made texts additionally influence the availability of the training data for the classifier. In our case, the training set for the Wikipedia sample was about a million word tokens (plus another quarter million punctuation marks).
Not all text generation tasks allow this large corpora, however, this is still the order of magnitude of a long novel (like classic Samuel Richardson's \textit{Clarissa}, with about 1.1 million tokens with punctuation) or several shorter ones. The frequency-based pipeline has been successfully tested before on two novels of joint size of under 60 thousand word tokens~\citep{ochab2024} and even shorter~\citep{argasinski2024}, three research papers yielding jointly 3400 tokens. 

In the subtask 1 of ``Voight-Kampff Generative AI Detection at PAN and ELOQUENT 2025''~\citep{bevendorff_overview_2025} (essays, news, and fiction genres as well as their obsfuscated versions) our pipeline \citep{ochab2025styloch} without hyperparameter optimisation has reached $F_1 = 0.823$ against the top result $F_1 = 0.898$.
A recent zero-shot detection solution achieved in \cite{sun2025zero} accuracy of 90,6\%. The average accuracy on three different structured texts datasets is 79,26\%. Both show that stylometric approach achieves better results.
In \cite{xu2024freqmark} FreqMark method was proposed for LLM generated text using frequency-based watermark. It shows robustness against paraphrasing and other attack methods. The accuracy of 98\% shows that not only stylometry-based methods perform well on paraphrased text. Stylometric methods were used in \cite{al2025arabic} for text fingerprints to detect text generated by LLMs. For four models: two Arabic and two general text, the accuracy vary for LLM models text generation detection between 88.23 and 98.07\% for social media content.


\begin{table*}[ht!]
    \centering
    \begin{tabular}{cccccccc}
        \toprule
        \textbf{Model} & \textbf{Falcon} & \textbf{GPT-3} & \textbf{GPT-4} & \textbf{LLaMA~2}	& \textbf{LLaMA~3} & \textbf{Orca} & \textbf{Human}\\
        \midrule
        \textbf{GPTZero prompt \#1} & 98\%	& 100\% & 96\% & 98\% & 98\% & 98\% & 100\% \\
        \textbf{GPTZero prompt \#2} & 100\% & 99\%	& 93\% & 95\% & 100\% & 99\% & N/A\\        
        \textbf{HIX prompt \#1} & 	3\%	& 5\% & 3\%	& 5\% & 4\%	& 0\% & 100\%\\                            
        \bottomrule
    \end{tabular}
    \caption{Commercial applications predictions}
    \label{tab:commercial}
\end{table*}

\subsection{Commercial applications}

We have tested two different commercial solutions. The test set was a randomly chosen set of 100 prompt results for each model separately. We also included 100 human-written text. The number of chosen samples is caused due to the high costs of each request of such tools. The results are presented in Table \ref{tab:commercial}. \cite{noauthor_gptzero_nodate} performs very well and predicts if the text is written by a human or a model almost perfectly. The disadvantage of this solution, similar to almost every commercial solution, is that it is a binary classification: human or AI. The models correctly identified the generated text as LLM generated, but there are no details on exactly what LLM was used for generation. Our proposed solution performs slightly worse than GPTZero, but it is a multi-label classifier. For comparison, we have tested the classifier developed by \cite{hix_ai_hix_nodate}. It performs very poorly and classifies almost every sample as written by humans.
\section{Further works}
\label{sec:further}

The results show that we can use stylometry for the English language to distinguish between LLMs and human written text. The next steps would be to perform the analysis in other languages, including low-resource languages.

The second way to extend this research is to use other stylometry libraries, classification methods, and more complex language models. Based on the results presented, the more complex models show that they are harder to differentiate from human written text compared to the less complex models. 

The third vector of further research is to extend the feature list with features encoding long memory and correlations in text, such as fractal-based features. As stylometry seems to be a good choice, there might be other ones that might be more precise.

Another extension can be the use of stylometry together with neural embeddings. A hybrid approach might increase the accuracy in generated text recognition.

Finally, the classification explanations obtained by different methods and from different classifiers should be verified for their consistency and stability across various domains.
\section*{Source code}
The data files and code used for text preprocessing and analysis can be found at \url{https://osf.io/dfz6k/}~\citep{ochab_repository_2025}.

The source code for text generation can be found in the repository:\\\url{https://github.com/kprzystalski/stylometry-llm}.\\
It includes: the URLs to the libraries, the code to get the data, preprocess it, and execute the experiment. It comes with setup guidelines, contains all parameters set for each model.

\section*{Acknowledgements}
We thank Tomasz Walkowiak for many insightful comments.

The research for this publication has been supported by a grant from the Priority Research Area DigiWorld under the Strategic Programme Excellence Initiative at Jagiellonian University.

JKO's research on the stylometric pipeline was financed by European Funds for Smart Economy, FENG program, CLARIN – Common Language Resources and Technology Infrastructure, project no.FENG.02.04-IP.040004/24-00.

This project has received funding from the European Union's Horizon 2020 research and innovation programme under grant agreement No 857533 and from the International Research Agendas Programme of the Foundation for Polish Science No MAB PLUS/2019/13.

The publication was created within the project of the Minister of Science and Higher Education “Support for the activity of Centers of Excellence established in Poland under Horizon 2020” on the basis of the contract number MEiN/2023/DIR/3796. 

\section*{Declaration of generative AI and AI-assisted technologies in the writing process}
 During the preparation of this work, the authors used GPT and Writefull LLM models to improve the readability, style and grammar of the text. After using this tool/service, the author(s) reviewed and edited the content as needed and take full responsibility for the content of the published article.

\bibliography{stylo}

\begin{thebibliography}{88}
\expandafter\ifx\csname natexlab\endcsname\relax\def\natexlab#1{#1}\fi
\providecommand{\url}[1]{\texttt{#1}}
\providecommand{\href}[2]{#2}
\providecommand{\path}[1]{#1}
\providecommand{\DOIprefix}{doi:}
\providecommand{\ArXivprefix}{arXiv:}
\providecommand{\URLprefix}{URL: }
\providecommand{\Pubmedprefix}{pmid:}
\providecommand{\doi}[1]{\href{http://dx.doi.org/#1}{\path{#1}}}
\providecommand{\Pubmed}[1]{\href{pmid:#1}{\path{#1}}}
\providecommand{\bibinfo}[2]{#2}
\ifx\xfnm\relax \def\xfnm[#1]{\unskip,\space#1}\fi
\bibitem[{noa(2025)}]{noauthor_gptzero_nodate}
 (\bibinfo{year}{2025}).
\newblock \bibinfo{title}{{GPTZero}}.
\newblock \URLprefix \url{https://gptzero.me/} \bibinfo{note}{{Accessed}: 2025-05-20}.
\bibitem[{Al-Shaibani \& Ahmed(2025)}]{al2025arabic}
\bibinfo{author}{Al-Shaibani, M.~S.}, \& \bibinfo{author}{Ahmed, M.} (\bibinfo{year}{2025}).
\newblock \bibinfo{title}{The arabic ai fingerprint: Stylometric analysis and detection of large language models text}.
\newblock {\it \bibinfo{journal}{arXiv preprint arXiv:2505.23276}\/}, .
\bibitem[{Almazrouei et~al.(2023)Almazrouei, Alobeidli, Alshamsi, Cappelli, Cojocaru, Debbah, Étienne Goffinet, Hesslow, Launay, Malartic, Mazzotta, Noune, Pannier \& Penedo}]{almazrouei2023falcon}
\bibinfo{author}{Almazrouei, E.}, \bibinfo{author}{Alobeidli, H.}, \bibinfo{author}{Alshamsi, A.}, \bibinfo{author}{Cappelli, A.}, \bibinfo{author}{Cojocaru, R.}, \bibinfo{author}{Debbah, M.}, \bibinfo{author}{Étienne Goffinet}, \bibinfo{author}{Hesslow, D.}, \bibinfo{author}{Launay, J.}, \bibinfo{author}{Malartic, Q.}, \bibinfo{author}{Mazzotta, D.}, \bibinfo{author}{Noune, B.}, \bibinfo{author}{Pannier, B.}, \& \bibinfo{author}{Penedo, G.} (\bibinfo{year}{2023}).
\newblock \bibinfo{title}{The falcon series of open language models}.
\newblock \href{http://arxiv.org/abs/2311.16867}{\tt arXiv:2311.16867}.
\bibitem[{Anand et~al.(2023)Anand, Nussbaum, Treat, Miller, Guo, Schmidt, Community, Duderstadt \& Mulyar}]{gpt4all}
\bibinfo{author}{Anand, Y.}, \bibinfo{author}{Nussbaum, Z.}, \bibinfo{author}{Treat, A.}, \bibinfo{author}{Miller, A.}, \bibinfo{author}{Guo, R.}, \bibinfo{author}{Schmidt, B.}, \bibinfo{author}{Community, G.}, \bibinfo{author}{Duderstadt, B.}, \& \bibinfo{author}{Mulyar, A.} (\bibinfo{year}{2023}).
\newblock \bibinfo{title}{Gpt4all: An ecosystem of open source compressed language models}.
\newblock \URLprefix \url{https://arxiv.org/abs/2311.04931}. \href{http://arxiv.org/abs/2311.04931}{\tt arXiv:2311.04931}.
\bibitem[{Argamon(2018)}]{argamon2018computational}
\bibinfo{author}{Argamon, S.} (\bibinfo{year}{2018}).
\newblock \bibinfo{title}{Computational forensic authorship analysis: Promises and pitfalls}.
\newblock {\it \bibinfo{journal}{Language and Law/Linguagem e Direito}\/},  {\it \bibinfo{volume}{5}\/}, \bibinfo{pages}{7--37}.
\bibitem[{Argasiński et~al.(2024)Argasiński, Grabska-Gradzińska, Przystalski, Ochab \& Walkowiak}]{argasinski2024}
\bibinfo{author}{Argasiński, J.~K.}, \bibinfo{author}{Grabska-Gradzińska, I.}, \bibinfo{author}{Przystalski, K.}, \bibinfo{author}{Ochab, J.~K.}, \& \bibinfo{author}{Walkowiak, T.} (\bibinfo{year}{2024}).
\newblock \bibinfo{title}{Stylometric analysis of large language model-generated commentaries in the context of medical neuroscience}.
\newblock {\it \bibinfo{journal}{International Conference …}\/},  (pp. \bibinfo{pages}{281--295}). \URLprefix \url{https://link.springer.com/chapter/10.1007/978-3-031-63775-9_20}. \DOIprefix\doi{10.1007/978-3-031-63775-9_20}.
\bibitem[{Bevendorff et~al.(2025)Bevendorff, Wang, Karlgren, Wiegmann, Tsivgun, Su, Xie, Abassy, Mansurov, Xing, Ta, Elozeiri, Gu, Tomar, Geng, Artemova, Shelmanov, Habash, Stamatatos, Gurevych, Nakov, Potthast \& Stein}]{bevendorff_overview_2025}
\bibinfo{author}{Bevendorff, J.}, \bibinfo{author}{Wang, Y.}, \bibinfo{author}{Karlgren, J.}, \bibinfo{author}{Wiegmann, M.}, \bibinfo{author}{Tsivgun, A.}, \bibinfo{author}{Su, J.}, \bibinfo{author}{Xie, Z.}, \bibinfo{author}{Abassy, M.}, \bibinfo{author}{Mansurov, J.}, \bibinfo{author}{Xing, R.}, \bibinfo{author}{Ta, M.~N.}, \bibinfo{author}{Elozeiri, K.~A.}, \bibinfo{author}{Gu, T.}, \bibinfo{author}{Tomar, R.~V.}, \bibinfo{author}{Geng, J.}, \bibinfo{author}{Artemova, E.}, \bibinfo{author}{Shelmanov, A.}, \bibinfo{author}{Habash, N.}, \bibinfo{author}{Stamatatos, E.}, \bibinfo{author}{Gurevych, I.}, \bibinfo{author}{Nakov, P.}, \bibinfo{author}{Potthast, M.}, \& \bibinfo{author}{Stein, B.} (\bibinfo{year}{2025}).
\newblock \bibinfo{title}{{Overview of the ``Voight-Kampff'' Generative AI Authorship Verification Task at PAN and ELOQUENT 2025}}.
\newblock In \bibinfo{editor}{G.~Faggioli}, \bibinfo{editor}{N.~Ferro}, \bibinfo{editor}{P.~Rosso}, \& \bibinfo{editor}{D.~Spina} (Eds.), {\it \bibinfo{booktitle}{Working Notes of CLEF 2025 -- Conference and Labs of the Evaluation Forum}\/} CEUR Workshop Proceedings.
\newblock \bibinfo{publisher}{CEUR-WS.org}.
\bibitem[{Bevendorff et~al.(2024)Bevendorff, Wiegmann, Karlgren, Dürlich, Gogoulou, Talman, Stamatatos, Potthast \& Stein}]{bevendorff_overview_2024}
\bibinfo{author}{Bevendorff, J.}, \bibinfo{author}{Wiegmann, M.}, \bibinfo{author}{Karlgren, J.}, \bibinfo{author}{Dürlich, L.}, \bibinfo{author}{Gogoulou, E.}, \bibinfo{author}{Talman, A.}, \bibinfo{author}{Stamatatos, E.}, \bibinfo{author}{Potthast, M.}, \& \bibinfo{author}{Stein, B.} (\bibinfo{year}{2024}).
\newblock \bibinfo{title}{Overview of the "{Voight}-{Kampff}" {Generative} {AI} {Authorship} {Verification} {Task} at {PAN} and {ELOQUENT} 2024}.
\newblock In \bibinfo{editor}{G.~Faggioli}, \bibinfo{editor}{N.~Ferro}, \bibinfo{editor}{P.~Galusc{'a}kov{'a}}, \& \bibinfo{editor}{A.~G. S.~d. Herrera} (Eds.), {\it \bibinfo{booktitle}{Working {Notes} of the {Conference} and {Labs} of the {Evaluation} {Forum} ({CLEF} 2024), {Grenoble}, {France}, 9-12 {September}, 2024}\/} (pp. \bibinfo{pages}{2486--2506}).
\newblock \bibinfo{publisher}{CEUR-WS.org} volume \bibinfo{volume}{3740} of {\it \bibinfo{series}{{CEUR} {Workshop} {Proceedings}}\/}.
\newblock \URLprefix \url{https://ceur-ws.org/Vol-3740/paper-225.pdf}.
\bibitem[{Bhat \& Parthasarathy(2020)}]{bhat_how_2020}
\bibinfo{author}{Bhat, M.~M.}, \& \bibinfo{author}{Parthasarathy, S.} (\bibinfo{year}{2020}).
\newblock \bibinfo{title}{How {Effectively} {Can} {Machines} {Defend} {Against} {Machine}-{Generated} {Fake} {News}? {An} {Empirical} {Study}}.
\newblock In \bibinfo{editor}{A.~Rogers}, \bibinfo{editor}{J.~Sedoc}, \& \bibinfo{editor}{A.~Rumshisky} (Eds.), {\it \bibinfo{booktitle}{Proceedings of the {First} {Workshop} on {Insights} from {Negative} {Results} in {NLP}}\/} (pp. \bibinfo{pages}{48--53}).
\newblock \bibinfo{address}{Online}: \bibinfo{publisher}{Association for Computational Linguistics}.
\newblock \URLprefix \url{https://aclanthology.org/2020.insights-1.7/}. \DOIprefix\doi{10.18653/v1/2020.insights-1.7}.
\bibitem[{Bhattacharjee et~al.(2023)Bhattacharjee, Kumarage, Moraffah \& Liu}]{bhattacharjee_conda_2023}
\bibinfo{author}{Bhattacharjee, A.}, \bibinfo{author}{Kumarage, T.}, \bibinfo{author}{Moraffah, R.}, \& \bibinfo{author}{Liu, H.} (\bibinfo{year}{2023}).
\newblock \bibinfo{title}{{ConDA}: {Contrastive} {Domain} {Adaptation} for {AI}-generated {Text} {Detection}}.
\newblock In \bibinfo{editor}{J.~C. Park}, \bibinfo{editor}{Y.~Arase}, \bibinfo{editor}{B.~Hu}, \bibinfo{editor}{W.~Lu}, \bibinfo{editor}{D.~Wijaya}, \bibinfo{editor}{A.~Purwarianti}, \& \bibinfo{editor}{A.~A. Krisnadhi} (Eds.), {\it \bibinfo{booktitle}{Proceedings of the 13th {International} {Joint} {Conference} on {Natural} {Language} {Processing} and the 3rd {Conference} of the {Asia}-{Pacific} {Chapter} of the {Association} for {Computational} {Linguistics} ({Volume} 1: {Long} {Papers})}\/} (pp. \bibinfo{pages}{598--610}).
\newblock \bibinfo{address}{Nusa Dua, Bali}: \bibinfo{publisher}{Association for Computational Linguistics}.
\newblock \URLprefix \url{https://aclanthology.org/2023.ijcnlp-main.40/}. \DOIprefix\doi{10.18653/v1/2023.ijcnlp-main.40}.
\bibitem[{Bozza et~al.(2023)Bozza, Roten, Jover, Cammarota, Pousaz \& Taroni}]{bozza2023model}
\bibinfo{author}{Bozza, S.}, \bibinfo{author}{Roten, C.-A.}, \bibinfo{author}{Jover, A.}, \bibinfo{author}{Cammarota, V.}, \bibinfo{author}{Pousaz, L.}, \& \bibinfo{author}{Taroni, F.} (\bibinfo{year}{2023}).
\newblock \bibinfo{title}{A model-independent redundancy measure for human versus chatgpt authorship discrimination using a bayesian probabilistic approach}.
\newblock {\it \bibinfo{journal}{Scientific Reports}\/},  {\it \bibinfo{volume}{13}\/}, \bibinfo{pages}{19217}.
\bibitem[{Brennan et~al.(2012)Brennan, Afroz \& Greenstadt}]{brennan2012adversarial}
\bibinfo{author}{Brennan, M.}, \bibinfo{author}{Afroz, S.}, \& \bibinfo{author}{Greenstadt, R.} (\bibinfo{year}{2012}).
\newblock \bibinfo{title}{Adversarial stylometry: Circumventing authorship recognition to preserve privacy and anonymity}.
\newblock {\it \bibinfo{journal}{ACM Transactions on Information and System Security (TISSEC)}\/},  {\it \bibinfo{volume}{15}\/}, \bibinfo{pages}{1--22}.
\bibitem[{Brooks et~al.(2024)Brooks, Eggert \& Peskoff}]{brooks_rise_2024}
\bibinfo{author}{Brooks, C.}, \bibinfo{author}{Eggert, S.}, \& \bibinfo{author}{Peskoff, D.} (\bibinfo{year}{2024}).
\newblock \bibinfo{title}{The {Rise} of {AI}-{Generated} {Content} in {Wikipedia}}.
\newblock In \bibinfo{editor}{L.~Lucie-Aimée}, \bibinfo{editor}{A.~Fan}, \bibinfo{editor}{T.~Gwadabe}, \bibinfo{editor}{I.~Johnson}, \bibinfo{editor}{F.~Petroni}, \& \bibinfo{editor}{D.~van Strien} (Eds.), {\it \bibinfo{booktitle}{Proceedings of the {First} {Workshop} on {Advancing} {Natural} {Language} {Processing} for {Wikipedia}}\/} (pp. \bibinfo{pages}{67--79}).
\newblock \bibinfo{address}{Miami, Florida, USA}: \bibinfo{publisher}{Association for Computational Linguistics}.
\newblock \URLprefix \url{https://aclanthology.org/2024.wikinlp-1.12/}. \DOIprefix\doi{10.18653/v1/2024.wikinlp-1.12}.
\bibitem[{Chen et~al.(2023)Chen, Kang, Zhai, Li, Singh \& Raj}]{chen_token_2023}
\bibinfo{author}{Chen, Y.}, \bibinfo{author}{Kang, H.}, \bibinfo{author}{Zhai, V.}, \bibinfo{author}{Li, L.}, \bibinfo{author}{Singh, R.}, \& \bibinfo{author}{Raj, B.} (\bibinfo{year}{2023}).
\newblock \bibinfo{title}{Token {Prediction} as {Implicit} {Classification} to {Identify} {LLM}-{Generated} {Text}}.
\newblock In \bibinfo{editor}{H.~Bouamor}, \bibinfo{editor}{J.~Pino}, \& \bibinfo{editor}{K.~Bali} (Eds.), {\it \bibinfo{booktitle}{Proceedings of the 2023 {Conference} on {Empirical} {Methods} in {Natural} {Language} {Processing}}\/} (pp. \bibinfo{pages}{13112--13120}).
\newblock \bibinfo{address}{Singapore}: \bibinfo{publisher}{Association for Computational Linguistics}.
\newblock \URLprefix \url{https://aclanthology.org/2023.emnlp-main.810/}. \DOIprefix\doi{10.18653/v1/2023.emnlp-main.810}.
\bibitem[{Crothers et~al.(2022)Crothers, Japkowicz, Viktor \& Branco}]{crothers_adversarial_2022}
\bibinfo{author}{Crothers, E.}, \bibinfo{author}{Japkowicz, N.}, \bibinfo{author}{Viktor, H.}, \& \bibinfo{author}{Branco, P.} (\bibinfo{year}{2022}).
\newblock \bibinfo{title}{Adversarial {Robustness} of {Neural}-{Statistical} {Features} in {Detection} of {Generative} {Transformers}}.
\newblock In {\it \bibinfo{booktitle}{2022 {International} {Joint} {Conference} on {Neural} {Networks} ({IJCNN})}\/} (pp. \bibinfo{pages}{1--8}).
\newblock \URLprefix \url{https://ieeexplore.ieee.org/document/9892269}. \DOIprefix\doi{10.1109/IJCNN55064.2022.9892269} \bibinfo{note}{iSSN: 2161-4407}.
\bibitem[{Crothers et~al.(2023)Crothers, Japkowicz \& Viktor}]{crothers_machine-generated_2023}
\bibinfo{author}{Crothers, E.}, \bibinfo{author}{Japkowicz, N.}, \& \bibinfo{author}{Viktor, H.~L.} (\bibinfo{year}{2023}).
\newblock \bibinfo{title}{Machine-{Generated} {Text}: {A} {Comprehensive} {Survey} of {Threat} {Models} and {Detection} {Methods}}.
\newblock {\it \bibinfo{journal}{IEEE Access}\/},  {\it \bibinfo{volume}{11}\/}, \bibinfo{pages}{70977--71002}. \URLprefix \url{https://doi.org/10.1109/ACCESS.2023.3294090}. \DOIprefix\doi{10.1109/ACCESS.2023.3294090}.
\bibitem[{Damodaran(2021)}]{damodaran_parrot_2021}
\bibinfo{author}{Damodaran, P.} (\bibinfo{year}{2021}).
\newblock \bibinfo{title}{Parrot: {Paraphrase} generation for {NLU}.}
\newblock \bibinfo{note}{Version Number: v1.0}.
\bibitem[{De~Angelis et~al.(2023)De~Angelis, Baglivo, Arzilli, Privitera, Ferragina, Tozzi \& Rizzo}]{de2023chatgpt}
\bibinfo{author}{De~Angelis, L.}, \bibinfo{author}{Baglivo, F.}, \bibinfo{author}{Arzilli, G.}, \bibinfo{author}{Privitera, G.~P.}, \bibinfo{author}{Ferragina, P.}, \bibinfo{author}{Tozzi, A.~E.}, \& \bibinfo{author}{Rizzo, C.} (\bibinfo{year}{2023}).
\newblock \bibinfo{title}{Chatgpt and the rise of large language models: the new ai-driven infodemic threat in public health}.
\newblock {\it \bibinfo{journal}{Frontiers in public health}\/},  {\it \bibinfo{volume}{11}\/}, \bibinfo{pages}{1166120}.
\bibitem[{Dhaini et~al.(2023)Dhaini, Poelman \& Erdogan}]{dhaini2023detecting}
\bibinfo{author}{Dhaini, M.}, \bibinfo{author}{Poelman, W.}, \& \bibinfo{author}{Erdogan, E.} (\bibinfo{year}{2023}).
\newblock \bibinfo{title}{Detecting chatgpt: A survey of the state of detecting chatgpt-generated text}.
\newblock {\it \bibinfo{journal}{arXiv preprint arXiv:2309.07689}\/}, .
\bibitem[{Ding et~al.(2017)Ding, Fung, Iqbal \& Cheung}]{ding2017learning}
\bibinfo{author}{Ding, S.~H.}, \bibinfo{author}{Fung, B.~C.}, \bibinfo{author}{Iqbal, F.}, \& \bibinfo{author}{Cheung, W.~K.} (\bibinfo{year}{2017}).
\newblock \bibinfo{title}{Learning stylometric representations for authorship analysis}.
\newblock {\it \bibinfo{journal}{IEEE transactions on cybernetics}\/},  {\it \bibinfo{volume}{49}\/}, \bibinfo{pages}{107--121}.
\bibitem[{Eder(2011)}]{eder_style-markers_2011}
\bibinfo{author}{Eder, M.} (\bibinfo{year}{2011}).
\newblock \bibinfo{title}{Style-{Markers} in {Authorship} {Attribution}: {A} {Cross}-{Language} {Study} of {The} {Authorial} {Fingerprint}}.
\newblock {\it \bibinfo{journal}{Studies in Polish Linguistics}\/},  (pp. \bibinfo{pages}{101--116}).
\bibitem[{Eder et~al.(2016)Eder, Kestemont \& Rybicki}]{eder2016RJStylometry}
\bibinfo{author}{Eder, M.}, \bibinfo{author}{Kestemont, M.}, \& \bibinfo{author}{Rybicki, J.} (\bibinfo{year}{2016}).
\newblock \bibinfo{title}{Stylometry with {R}: {A} {Package} for {Computational} {Text} {Analysis}}.
\newblock {\it \bibinfo{journal}{The R Journal}\/},  {\it \bibinfo{volume}{8}\/}, \bibinfo{pages}{1--15}. \DOIprefix\doi{10.32614/RJ-2016-007}.
\bibitem[{Evert et~al.(2017)Evert, Proisl, Jannidis, Reger, Pielström, Schöch \& Vitt}]{evert_understanding_2017}
\bibinfo{author}{Evert, S.}, \bibinfo{author}{Proisl, T.}, \bibinfo{author}{Jannidis, F.}, \bibinfo{author}{Reger, I.}, \bibinfo{author}{Pielström, S.}, \bibinfo{author}{Schöch, C.}, \& \bibinfo{author}{Vitt, T.} (\bibinfo{year}{2017}).
\newblock \bibinfo{title}{Understanding and explaining {Delta} measures for authorship attribution}.
\newblock {\it \bibinfo{journal}{Digital Scholarship in the Humanities}\/},  {\it \bibinfo{volume}{32}\/}, \bibinfo{pages}{ii4--ii16}. \URLprefix \url{http://academic.oup.com/dsh/article/32/suppl_2/ii4/3865676}. \DOIprefix\doi{10.1093/llc/fqx023}.
\bibitem[{Fagni et~al.(2021)Fagni, Falchi, Gambini, Martella \& Tesconi}]{fagni_tweepfake_2021}
\bibinfo{author}{Fagni, T.}, \bibinfo{author}{Falchi, F.}, \bibinfo{author}{Gambini, M.}, \bibinfo{author}{Martella, A.}, \& \bibinfo{author}{Tesconi, M.} (\bibinfo{year}{2021}).
\newblock \bibinfo{title}{{TweepFake}: {About} detecting deepfake tweets}.
\newblock {\it \bibinfo{journal}{PLOS ONE}\/},  {\it \bibinfo{volume}{16}\/}, \bibinfo{pages}{e0251415}. \URLprefix \url{https://journals.plos.org/plosone/article?id=10.1371/journal.pone.0251415}. \DOIprefix\doi{10.1371/journal.pone.0251415}.
\newblock \bibinfo{note}{Publisher: Public Library of Science}.
\bibitem[{Geng et~al.(2020)Geng, Huang \& Chen}]{geng_recent_2020}
\bibinfo{author}{Geng, C.}, \bibinfo{author}{Huang, S.-j.}, \& \bibinfo{author}{Chen, S.} (\bibinfo{year}{2020}).
\newblock \bibinfo{title}{Recent advances in open set recognition: {A} survey}.
\newblock {\it \bibinfo{journal}{IEEE transactions on pattern analysis and machine intelligence}\/},  {\it \bibinfo{volume}{43}\/}, \bibinfo{pages}{3614--3631}.
\newblock \bibinfo{note}{Publisher: IEEE}.
\bibitem[{Guo et~al.(2023)Guo, Zhang, Wang, Jiang, Nie, Ding, Yue \& Wu}]{guo_how_2023}
\bibinfo{author}{Guo, B.}, \bibinfo{author}{Zhang, X.}, \bibinfo{author}{Wang, Z.}, \bibinfo{author}{Jiang, M.}, \bibinfo{author}{Nie, J.}, \bibinfo{author}{Ding, Y.}, \bibinfo{author}{Yue, J.}, \& \bibinfo{author}{Wu, Y.} (\bibinfo{year}{2023}).
\newblock \bibinfo{title}{How {Close} is {ChatGPT} to {Human} {Experts}? {Comparison} {Corpus}, {Evaluation}, and {Detection}}.
\newblock {\it \bibinfo{journal}{CoRR}\/},  {\it \bibinfo{volume}{abs/2301.07597}\/}. \URLprefix \url{https://doi.org/10.48550/arXiv.2301.07597}. \DOIprefix\doi{10.48550/ARXIV.2301.07597}.
\newblock \bibinfo{note}{ArXiv: 2301.07597}.
\bibitem[{Guo et~al.(2024{\natexlab{a}})Guo, Yang, Ma \& Ruan}]{guo_blgav_2024}
\bibinfo{author}{Guo, L.}, \bibinfo{author}{Yang, W.}, \bibinfo{author}{Ma, L.}, \& \bibinfo{author}{Ruan, J.} (\bibinfo{year}{2024}{\natexlab{a}}).
\newblock \bibinfo{title}{{BLGAV}: {Generative} {AI} {Author} {Verification} {Model} {Based} on {BERT} and {BiLSTM}}.
\newblock In \bibinfo{editor}{G.~Faggioli}, \bibinfo{editor}{N.~Ferro}, \bibinfo{editor}{P.~Galušč{'a}kov{'a}}, \& \bibinfo{editor}{A.~G.~S. Herrera} (Eds.), {\it \bibinfo{booktitle}{Working {Notes} {Papers} of the {CLEF} 2024 {Evaluation} {Labs}}\/} (pp. \bibinfo{pages}{2585--2592}).
\newblock \bibinfo{publisher}{CEUR-WS.org}.
\newblock \URLprefix \url{http://ceur-ws.org/Vol-3740/paper-237.pdf}.
\bibitem[{Guo et~al.(2024{\natexlab{b}})Guo, Han, Chen \& Peng}]{guo_machine-generated_2024}
\bibinfo{author}{Guo, M.}, \bibinfo{author}{Han, Z.}, \bibinfo{author}{Chen, H.}, \& \bibinfo{author}{Peng, J.} (\bibinfo{year}{2024}{\natexlab{b}}).
\newblock \bibinfo{title}{A {Machine}-{Generated} {Text} {Detection} {Model} {Based} on {Text} {Multi}-{Feature} {Fusion}}.
\newblock In \bibinfo{editor}{G.~Faggioli}, \bibinfo{editor}{N.~Ferro}, \bibinfo{editor}{P.~Galusc{'a}kov{'a}}, \& \bibinfo{editor}{A.~G. S.~d. Herrera} (Eds.), {\it \bibinfo{booktitle}{Working {Notes} of the {Conference} and {Labs} of the {Evaluation} {Forum} ({CLEF} 2024), {Grenoble}, {France}, 9-12 {September}, 2024}\/} (pp. \bibinfo{pages}{2593--2602}).
\newblock \bibinfo{publisher}{CEUR-WS.org} volume \bibinfo{volume}{3740} of {\it \bibinfo{series}{{CEUR} {Workshop} {Proceedings}}\/}.
\newblock \URLprefix \url{https://ceur-ws.org/Vol-3740/paper-238.pdf}.
\bibitem[{Hicke \& Mimno(2023)}]{hicke2023t5}
\bibinfo{author}{Hicke, R.}, \& \bibinfo{author}{Mimno, D.} (\bibinfo{year}{2023}).
\newblock \bibinfo{title}{T5 meets tybalt: Author attribution in early modern english drama using large language models}.
\newblock {\it \bibinfo{journal}{arXiv preprint arXiv:2310.18454}\/}, .
\bibitem[{{HIX A.I.}(2025)}]{hix_ai_hix_nodate}
\bibinfo{author}{{HIX A.I.}} (\bibinfo{year}{2025}).
\newblock \bibinfo{title}{{HIX} {AI} {Detector}}.
\newblock \URLprefix \url{https://bypass.hix.ai/ai-detector} \bibinfo{note}{{Accessed}: 2025-05-20}.
\bibitem[{Hu et~al.(2023)Hu, Ou, Acharya, Ding, D'Gama \& ...}]{hu2023tdrlm}
\bibinfo{author}{Hu, X.}, \bibinfo{author}{Ou, W.}, \bibinfo{author}{Acharya, S.}, \bibinfo{author}{Ding, S.}, \bibinfo{author}{D'Gama, R.}, \& \bibinfo{author}{...} (\bibinfo{year}{2023}).
\newblock \bibinfo{title}{Tdrlm: Stylometric learning for authorship verification by topic-debiasing}.
\newblock {\it \bibinfo{journal}{Expert Systems with Applications}\/}, .
\bibitem[{Huang et~al.(2024{\natexlab{a}})Huang, Chen \& Shu}]{huang2024can}
\bibinfo{author}{Huang, B.}, \bibinfo{author}{Chen, C.}, \& \bibinfo{author}{Shu, K.} (\bibinfo{year}{2024}{\natexlab{a}}).
\newblock \bibinfo{title}{Can large language models identify authorship?}
\newblock {\it \bibinfo{journal}{arXiv preprint arXiv:2403.08213}\/}, .
\bibitem[{Huang et~al.(2024{\natexlab{b}})Huang, Ruan, Huang, Jin, Dong, Wu, Bensalem, Mu, Qi, Zhao et~al.}]{huang2024survey}
\bibinfo{author}{Huang, X.}, \bibinfo{author}{Ruan, W.}, \bibinfo{author}{Huang, W.}, \bibinfo{author}{Jin, G.}, \bibinfo{author}{Dong, Y.}, \bibinfo{author}{Wu, C.}, \bibinfo{author}{Bensalem, S.}, \bibinfo{author}{Mu, R.}, \bibinfo{author}{Qi, Y.}, \bibinfo{author}{Zhao, X.} et~al. (\bibinfo{year}{2024}{\natexlab{b}}).
\newblock \bibinfo{title}{A survey of safety and trustworthiness of large language models through the lens of verification and validation}.
\newblock {\it \bibinfo{journal}{Artificial Intelligence Review}\/},  {\it \bibinfo{volume}{57}\/}, \bibinfo{pages}{175}.
\bibitem[{Hung et~al.(2023)Hung, Hu, Hu \& Lee}]{hung2023wrote}
\bibinfo{author}{Hung, C.-Y.}, \bibinfo{author}{Hu, Z.}, \bibinfo{author}{Hu, Y.}, \& \bibinfo{author}{Lee, R. K.-W.} (\bibinfo{year}{2023}).
\newblock \bibinfo{title}{Who wrote it and why? prompting large-language models for authorship verification}.
\newblock {\it \bibinfo{journal}{arXiv preprint arXiv:2310.08123}\/}, .
\bibitem[{Ke et~al.(2017)Ke, Meng, Finley, Wang, Chen, Ma, Ye \& Liu}]{ke2017lightgbm}
\bibinfo{author}{Ke, G.}, \bibinfo{author}{Meng, Q.}, \bibinfo{author}{Finley, T.}, \bibinfo{author}{Wang, T.}, \bibinfo{author}{Chen, W.}, \bibinfo{author}{Ma, W.}, \bibinfo{author}{Ye, Q.}, \& \bibinfo{author}{Liu, T.-Y.} (\bibinfo{year}{2017}).
\newblock \bibinfo{title}{Lightgbm: A highly efficient gradient boosting decision tree}.
\newblock {\it \bibinfo{journal}{Advances in neural information processing systems}\/},  {\it \bibinfo{volume}{30}\/}, \bibinfo{pages}{3146--3154}.
\bibitem[{Krishna et~al.(2023)Krishna, Song, Karpinska, Wieting \& Iyyer}]{krishna_paraphrasing_2023}
\bibinfo{author}{Krishna, K.}, \bibinfo{author}{Song, Y.}, \bibinfo{author}{Karpinska, M.}, \bibinfo{author}{Wieting, J.}, \& \bibinfo{author}{Iyyer, M.} (\bibinfo{year}{2023}).
\newblock \bibinfo{title}{Paraphrasing evades detectors of {AI}-generated text, but retrieval is an effective defense}.
\newblock {\it \bibinfo{journal}{arXiv preprint arXiv:2303.13408}\/}, .
\bibitem[{Kumarage et~al.(2023)Kumarage, Garland, Bhattacharjee, Trapeznikov, Ruston \& Liu}]{kumarage_stylometric_2023}
\bibinfo{author}{Kumarage, T.}, \bibinfo{author}{Garland, J.}, \bibinfo{author}{Bhattacharjee, A.}, \bibinfo{author}{Trapeznikov, K.}, \bibinfo{author}{Ruston, S.~W.}, \& \bibinfo{author}{Liu, H.} (\bibinfo{year}{2023}).
\newblock \bibinfo{title}{Stylometric {Detection} of {AI}-{Generated} {Text} in {Twitter} {Timelines}}.
\newblock {\it \bibinfo{journal}{CoRR}\/},  {\it \bibinfo{volume}{abs/2303.03697}\/}. \URLprefix \url{https://doi.org/10.48550/arXiv.2303.03697}. \DOIprefix\doi{10.48550/ARXIV.2303.03697}.
\newblock \bibinfo{note}{ArXiv: 2303.03697}.
\bibitem[{Kumarage \& Liu(2023)}]{kumarage2023neural}
\bibinfo{author}{Kumarage, T.}, \& \bibinfo{author}{Liu, H.} (\bibinfo{year}{2023}).
\newblock \bibinfo{title}{Neural authorship attribution: Stylometric analysis on large language models}.
\newblock In {\it \bibinfo{booktitle}{2023 International Conference on Cyber-Enabled Distributed Computing and Knowledge Discovery (CyberC)}\/} (pp. \bibinfo{pages}{51--54}).
\newblock \bibinfo{organization}{IEEE}.
\bibitem[{Lewis et~al.(2019)Lewis, Liu, Goyal, Ghazvininejad, Mohamed, Levy, Stoyanov \& Zettlemoyer}]{bartsum}
\bibinfo{author}{Lewis, M.}, \bibinfo{author}{Liu, Y.}, \bibinfo{author}{Goyal, N.}, \bibinfo{author}{Ghazvininejad, M.}, \bibinfo{author}{Mohamed, A.}, \bibinfo{author}{Levy, O.}, \bibinfo{author}{Stoyanov, V.}, \& \bibinfo{author}{Zettlemoyer, L.} (\bibinfo{year}{2019}).
\newblock \bibinfo{title}{Bart: Denoising sequence-to-sequence pre-training for natural language generation, translation, and comprehension}.
\newblock {\it \bibinfo{journal}{arXiv preprint arXiv:1910.13461}\/}, .
\bibitem[{Lhoest et~al.(2021)Lhoest, Villanova~del Moral, Jernite, Thakur, von Platen, Patil, Chaumond, Drame, Plu, Tunstall, Davison, {\v{S}}a{\v{s}}ko, Chhablani, Malik, Brandeis, Le~Scao, Sanh, Xu, Patry, McMillan-Major, Schmid, Gugger, Delangue, Matussi{\`e}re, Debut, Bekman, Cistac, Goehringer, Mustar, Lagunas, Rush \& Wolf}]{lhoest-etal-2021-datasets}
\bibinfo{author}{Lhoest, Q.}, \bibinfo{author}{Villanova~del Moral, A.}, \bibinfo{author}{Jernite, Y.}, \bibinfo{author}{Thakur, A.}, \bibinfo{author}{von Platen, P.}, \bibinfo{author}{Patil, S.}, \bibinfo{author}{Chaumond, J.}, \bibinfo{author}{Drame, M.}, \bibinfo{author}{Plu, J.}, \bibinfo{author}{Tunstall, L.}, \bibinfo{author}{Davison, J.}, \bibinfo{author}{{\v{S}}a{\v{s}}ko, M.}, \bibinfo{author}{Chhablani, G.}, \bibinfo{author}{Malik, B.}, \bibinfo{author}{Brandeis, S.}, \bibinfo{author}{Le~Scao, T.}, \bibinfo{author}{Sanh, V.}, \bibinfo{author}{Xu, C.}, \bibinfo{author}{Patry, N.}, \bibinfo{author}{McMillan-Major, A.}, \bibinfo{author}{Schmid, P.}, \bibinfo{author}{Gugger, S.}, \bibinfo{author}{Delangue, C.}, \bibinfo{author}{Matussi{\`e}re, T.}, \bibinfo{author}{Debut, L.}, \bibinfo{author}{Bekman, S.}, \bibinfo{author}{Cistac, P.}, \bibinfo{author}{Goehringer, T.}, \bibinfo{author}{Mustar, V.}, \bibinfo{author}{Lagunas, F.}, \bibinfo{author}{Rush, A.}, \& \bibinfo{author}{Wolf, T.}
  (\bibinfo{year}{2021}).
\newblock \bibinfo{title}{Datasets: A community library for natural language processing}.
\newblock In {\it \bibinfo{booktitle}{Proceedings of the 2021 Conference on Empirical Methods in Natural Language Processing: System Demonstrations}\/} (pp. \bibinfo{pages}{175--184}).
\newblock \bibinfo{address}{Online and Punta Cana, Dominican Republic}: \bibinfo{publisher}{Association for Computational Linguistics}.
\newblock \URLprefix \url{https://aclanthology.org/2021.emnlp-demo.21}. \href{http://arxiv.org/abs/2109.02846}{\tt arXiv:2109.02846}.
\bibitem[{Li et~al.(2024{\natexlab{a}})Li, Li, Cui, Bi, Wang, Wang, Yang, Shi \& Zhang}]{li_mage_2024}
\bibinfo{author}{Li, Y.}, \bibinfo{author}{Li, Q.}, \bibinfo{author}{Cui, L.}, \bibinfo{author}{Bi, W.}, \bibinfo{author}{Wang, Z.}, \bibinfo{author}{Wang, L.}, \bibinfo{author}{Yang, L.}, \bibinfo{author}{Shi, S.}, \& \bibinfo{author}{Zhang, Y.} (\bibinfo{year}{2024}{\natexlab{a}}).
\newblock \bibinfo{title}{{MAGE}: {Machine}-generated {Text} {Detection} in the {Wild}}.
\newblock In \bibinfo{editor}{L.-W. Ku}, \bibinfo{editor}{A.~Martins}, \& \bibinfo{editor}{V.~Srikumar} (Eds.), {\it \bibinfo{booktitle}{Proceedings of the 62nd {Annual} {Meeting} of the {Association} for {Computational} {Linguistics} ({Volume} 1: {Long} {Papers})}\/} (pp. \bibinfo{pages}{36--53}).
\newblock \bibinfo{address}{Bangkok, Thailand}: \bibinfo{publisher}{Association for Computational Linguistics}.
\newblock \URLprefix \url{https://aclanthology.org/2024.acl-long.3/}. \DOIprefix\doi{10.18653/v1/2024.acl-long.3}.
\bibitem[{Li et~al.(2024{\natexlab{b}})Li, Shi, Liu, Yang, Payani, Liu \& Du}]{li_quantifying_2024}
\bibinfo{author}{Li, Z.}, \bibinfo{author}{Shi, Y.}, \bibinfo{author}{Liu, Z.}, \bibinfo{author}{Yang, F.}, \bibinfo{author}{Payani, A.}, \bibinfo{author}{Liu, N.}, \& \bibinfo{author}{Du, M.} (\bibinfo{year}{2024}{\natexlab{b}}).
\newblock \bibinfo{title}{Quantifying {Multilingual} {Performance} of {Large} {Language} {Models} {Across} {Languages}}.
\newblock \URLprefix \url{https://arxiv.org/abs/2404.11553}. \DOIprefix\doi{10.48550/ARXIV.2404.11553} \bibinfo{note}{version Number: 2}.
\bibitem[{Liu et~al.(2024{\natexlab{a}})Liu, Diab \& Fried}]{liu_evaluating_2024}
\bibinfo{author}{Liu, A.}, \bibinfo{author}{Diab, M.}, \& \bibinfo{author}{Fried, D.} (\bibinfo{year}{2024}{\natexlab{a}}).
\newblock \bibinfo{title}{Evaluating {Large} {Language} {Model} {Biases} in {Persona}-{Steered} {Generation}}.
\newblock \URLprefix \url{http://arxiv.org/abs/2405.20253}. \DOIprefix\doi{10.48550/arXiv.2405.20253} \bibinfo{note}{arXiv:2405.20253 [cs]}.
\bibitem[{Liu et~al.(2023{\natexlab{a}})Liu, Han, Ma, Zhang, Yang, Tian \& ...}]{gpt4}
\bibinfo{author}{Liu, Y.}, \bibinfo{author}{Han, T.}, \bibinfo{author}{Ma, S.}, \bibinfo{author}{Zhang, J.}, \bibinfo{author}{Yang, Y.}, \bibinfo{author}{Tian, J.}, \& \bibinfo{author}{...} (\bibinfo{year}{2023}{\natexlab{a}}).
\newblock \bibinfo{title}{Summary of chatgpt/gpt-4 research and perspective towards the future of large language models}.
\newblock {\it \bibinfo{journal}{arXiv preprint arXiv …}\/}, .
\bibitem[{Liu et~al.(2023{\natexlab{b}})Liu, Zhang, Zhang, Yue, Zhao, Cheng, Zhang \& Hu}]{liu_argugpt_2023}
\bibinfo{author}{Liu, Y.}, \bibinfo{author}{Zhang, Z.}, \bibinfo{author}{Zhang, W.}, \bibinfo{author}{Yue, S.}, \bibinfo{author}{Zhao, X.}, \bibinfo{author}{Cheng, X.}, \bibinfo{author}{Zhang, Y.}, \& \bibinfo{author}{Hu, H.} (\bibinfo{year}{2023}{\natexlab{b}}).
\newblock \bibinfo{title}{{ArguGPT}: evaluating, understanding and identifying argumentative essays generated by {GPT} models}.
\newblock {\it \bibinfo{journal}{CoRR}\/},  {\it \bibinfo{volume}{abs/2304.07666}\/}. \URLprefix \url{https://doi.org/10.48550/arXiv.2304.07666}. \DOIprefix\doi{10.48550/ARXIV.2304.07666}.
\newblock \bibinfo{note}{ArXiv: 2304.07666}.
\bibitem[{Liu et~al.(2024{\natexlab{b}})Liu, Yao, Li \& Luo}]{liu_detectability_2024}
\bibinfo{author}{Liu, Z.}, \bibinfo{author}{Yao, Z.}, \bibinfo{author}{Li, F.}, \& \bibinfo{author}{Luo, B.} (\bibinfo{year}{2024}{\natexlab{b}}).
\newblock \bibinfo{title}{On the {Detectability} of {ChatGPT} {Content}: {Benchmarking}, {Methodology}, and {Evaluation} through the {Lens} of {Academic} {Writing}}.
\newblock \URLprefix \url{http://arxiv.org/abs/2306.05524}. \DOIprefix\doi{10.48550/arXiv.2306.05524} \bibinfo{note}{arXiv:2306.05524 [cs]}.
\bibitem[{Lorenz et~al.(2024)Lorenz, Aygüler, Schlatt \& Mirzakhmedova}]{lorenz_baselineavengers_2024}
\bibinfo{author}{Lorenz, L.}, \bibinfo{author}{Aygüler, F.~Z.}, \bibinfo{author}{Schlatt, F.}, \& \bibinfo{author}{Mirzakhmedova, N.} (\bibinfo{year}{2024}).
\newblock \bibinfo{title}{{BaselineAvengers} at {PAN} 2024: {Often}-{Forgotten} {Baselines} for {LLM}-{Generated} {Text} {Detection}}.
\newblock In \bibinfo{editor}{G.~Faggioli}, \bibinfo{editor}{N.~Ferro}, \bibinfo{editor}{P.~Galusc{'a}kov{'a}}, \& \bibinfo{editor}{A.~G. S.~d. Herrera} (Eds.), {\it \bibinfo{booktitle}{Working {Notes} of the {Conference} and {Labs} of the {Evaluation} {Forum} ({CLEF} 2024), {Grenoble}, {France}, 9-12 {September}, 2024}\/} (pp. \bibinfo{pages}{2761--2768}).
\newblock \bibinfo{publisher}{CEUR-WS.org} volume \bibinfo{volume}{3740} of {\it \bibinfo{series}{{CEUR} {Workshop} {Proceedings}}\/}.
\newblock \URLprefix \url{https://ceur-ws.org/Vol-3740/paper-262.pdf}.
\bibitem[{Lu et~al.(2024)Lu, Liu, He, Ong, Wang \& Tang}]{lu_large_2024}
\bibinfo{author}{Lu, N.}, \bibinfo{author}{Liu, S.}, \bibinfo{author}{He, R.}, \bibinfo{author}{Ong, Y.-S.}, \bibinfo{author}{Wang, Q.}, \& \bibinfo{author}{Tang, K.} (\bibinfo{year}{2024}).
\newblock \bibinfo{title}{Large {Language} {Models} can be {Guided} to {Evade} {AI}-generated {Text} {Detection}}.
\newblock {\it \bibinfo{journal}{Transactions on Machine Learning Research}\/},  {\it \bibinfo{volume}{2024}\/}. \URLprefix \url{https://openreview.net/forum?id=lLE0mWzUrr}.
\bibitem[{Lund et~al.(2023)Lund, Wang, Mannuru, Nie, Shimray \& Wang}]{lund2023chatgpt}
\bibinfo{author}{Lund, B.~D.}, \bibinfo{author}{Wang, T.}, \bibinfo{author}{Mannuru, N.~R.}, \bibinfo{author}{Nie, B.}, \bibinfo{author}{Shimray, S.}, \& \bibinfo{author}{Wang, Z.} (\bibinfo{year}{2023}).
\newblock \bibinfo{title}{Chatgpt and a new academic reality: Artificial intelligence-written research papers and the ethics of the large language models in scholarly publishing}.
\newblock {\it \bibinfo{journal}{Journal of the Association for Information Science and Technology}\/},  {\it \bibinfo{volume}{74}\/}, \bibinfo{pages}{570--581}.
\bibitem[{Lundberg et~al.(2020)Lundberg, Erion, Chen, DeGrave, Prutkin, Nair, Katz, Himmelfarb, Bansal \& Lee}]{lundberg2020local2global}
\bibinfo{author}{Lundberg, S.~M.}, \bibinfo{author}{Erion, G.}, \bibinfo{author}{Chen, H.}, \bibinfo{author}{DeGrave, A.}, \bibinfo{author}{Prutkin, J.~M.}, \bibinfo{author}{Nair, B.}, \bibinfo{author}{Katz, R.}, \bibinfo{author}{Himmelfarb, J.}, \bibinfo{author}{Bansal, N.}, \& \bibinfo{author}{Lee, S.-I.} (\bibinfo{year}{2020}).
\newblock \bibinfo{title}{From local explanations to global understanding with explainable ai for trees}.
\newblock {\it \bibinfo{journal}{Nature Machine Intelligence}\/},  {\it \bibinfo{volume}{2}\/}, \bibinfo{pages}{2522--5839}.
\bibitem[{Lüdeling \& Kytö(2008)}]{ludeling_corpus_2008}
\bibinfo{editor}{Lüdeling, A.}, \& \bibinfo{editor}{Kytö, M.} (Eds.) (\bibinfo{year}{2008}).
\newblock {\it \bibinfo{title}{Corpus linguistics: an international handbook}\/} volume~\bibinfo{volume}{1} of {\it \bibinfo{series}{Handbücher zur {Sprach}- und {Kommunikationswissenschaft}}\/}.
\newblock \bibinfo{address}{Berlin New York}: \bibinfo{publisher}{Walter de Gruyter}.
\bibitem[{Macko et~al.(2023)Macko, Moro, Uchendu, Lucas, Yamashita, Pikuliak, Srba, Le, Lee, Simko \& Bielikova}]{macko_multitude_2023}
\bibinfo{author}{Macko, D.}, \bibinfo{author}{Moro, R.}, \bibinfo{author}{Uchendu, A.}, \bibinfo{author}{Lucas, J.}, \bibinfo{author}{Yamashita, M.}, \bibinfo{author}{Pikuliak, M.}, \bibinfo{author}{Srba, I.}, \bibinfo{author}{Le, T.}, \bibinfo{author}{Lee, D.}, \bibinfo{author}{Simko, J.}, \& \bibinfo{author}{Bielikova, M.} (\bibinfo{year}{2023}).
\newblock \bibinfo{title}{{MULTITuDE}: {Large}-{Scale} {Multilingual} {Machine}-{Generated} {Text} {Detection} {Benchmark}}.
\newblock In \bibinfo{editor}{H.~Bouamor}, \bibinfo{editor}{J.~Pino}, \& \bibinfo{editor}{K.~Bali} (Eds.), {\it \bibinfo{booktitle}{Proceedings of the 2023 {Conference} on {Empirical} {Methods} in {Natural} {Language} {Processing}}\/} (pp. \bibinfo{pages}{9960--9987}).
\newblock \bibinfo{address}{Singapore}: \bibinfo{publisher}{Association for Computational Linguistics}.
\newblock \URLprefix \url{https://aclanthology.org/2023.emnlp-main.616/}. \DOIprefix\doi{10.18653/v1/2023.emnlp-main.616}.
\bibitem[{Mikros et~al.(2023)Mikros, Koursaris, Bilianos \& ...}]{mikros2023transformers}
\bibinfo{author}{Mikros, G.}, \bibinfo{author}{Koursaris, A.}, \bibinfo{author}{Bilianos, D.}, \& \bibinfo{author}{...} (\bibinfo{year}{2023}).
\newblock \bibinfo{title}{Ai-writing detection using an ensemble of transformers and stylometric features.}
\newblock {\it \bibinfo{journal}{IberLEF …}\/}, .
\bibitem[{Miralles et~al.(2024)Miralles, Martín \& Camacho}]{miralles_team_2024}
\bibinfo{author}{Miralles, P.}, \bibinfo{author}{Martín, A.}, \& \bibinfo{author}{Camacho, D.} (\bibinfo{year}{2024}).
\newblock \bibinfo{title}{Team aida at {PAN}: {Ensembling} {Normalized} {Log} {Probabilities}}.
\newblock In \bibinfo{editor}{G.~Faggioli}, \bibinfo{editor}{N.~Ferro}, \bibinfo{editor}{P.~Galušč{'a}kov{'a}}, \& \bibinfo{editor}{A.~G.~S. Herrera} (Eds.), {\it \bibinfo{booktitle}{Working {Notes} {Papers} of the {CLEF} 2024 {Evaluation} {Labs}}\/} (pp. \bibinfo{pages}{2807--2813}).
\newblock \bibinfo{publisher}{CEUR-WS.org}.
\newblock \URLprefix \url{http://ceur-ws.org/Vol-3740/paper-268.pdf}.
\bibitem[{Montani et~al.(2023)Montani, Honnibal, Honnibal, Boyd, Landeghem \& Peters}]{ines_montani_2023_10009823}
\bibinfo{author}{Montani, I.}, \bibinfo{author}{Honnibal, M.}, \bibinfo{author}{Honnibal, M.}, \bibinfo{author}{Boyd, A.}, \bibinfo{author}{Landeghem, S.~V.}, \& \bibinfo{author}{Peters, H.} (\bibinfo{year}{2023}).
\newblock \bibinfo{title}{{explosion/spaCy: v3.7.2: Fixes for APIs and requirements}}.
\newblock \URLprefix \url{https://doi.org/10.5281/zenodo.10009823}. \DOIprefix\doi{10.5281/zenodo.10009823}.
\bibitem[{Mosteller(1968)}]{mosteller1968association}
\bibinfo{author}{Mosteller, F.} (\bibinfo{year}{1968}).
\newblock \bibinfo{title}{Association and estimation in contingency tables}.
\newblock {\it \bibinfo{journal}{Journal of the American Statistical Association}\/},  {\it \bibinfo{volume}{63}\/}, \bibinfo{pages}{1--28}.
\bibitem[{Mukherjee et~al.(2023)Mukherjee, Mitra, Jawahar, Agarwal, Palangi \& Awadallah}]{orca}
\bibinfo{author}{Mukherjee, S.}, \bibinfo{author}{Mitra, A.}, \bibinfo{author}{Jawahar, G.}, \bibinfo{author}{Agarwal, S.}, \bibinfo{author}{Palangi, H.}, \& \bibinfo{author}{Awadallah, A.} (\bibinfo{year}{2023}).
\newblock \bibinfo{title}{Orca: Progressive learning from complex explanation traces of gpt-4}.
\newblock {\it \bibinfo{journal}{arXiv preprint arXiv:2306.02707}\/}, .
\bibitem[{Neal et~al.(2017)Neal, Sundararajan, Fatima, Yan, Xiang \& Woodard}]{neal2017surveying}
\bibinfo{author}{Neal, T.}, \bibinfo{author}{Sundararajan, K.}, \bibinfo{author}{Fatima, A.}, \bibinfo{author}{Yan, Y.}, \bibinfo{author}{Xiang, Y.}, \& \bibinfo{author}{Woodard, D.} (\bibinfo{year}{2017}).
\newblock \bibinfo{title}{Surveying stylometry techniques and applications}.
\newblock {\it \bibinfo{journal}{ACM Computing Surveys (CSuR)}\/},  {\it \bibinfo{volume}{50}\/}, \bibinfo{pages}{1--36}.
\bibitem[{Ochab et~al.(2025{\natexlab{a}})Ochab, Argasiński, Grabska-Gradzińska \& Przystalski}]{ochab_repository_2025}
\bibinfo{author}{Ochab, J.~K.}, \bibinfo{author}{Argasiński, J.}, \bibinfo{author}{Grabska-Gradzińska, I.}, \& \bibinfo{author}{Przystalski, K.} (\bibinfo{year}{2025}{\natexlab{a}}).
\newblock \bibinfo{title}{Repository for: {Stylometry} recognizes human and {LLM}-generated texts in short samples}, .
\newblock \URLprefix \url{https://osf.io/dfz6k/}. \DOIprefix\doi{10.17605/OSF.IO/DFZ6K}.
\newblock \bibinfo{note}{Publisher: OSF}.
\bibitem[{Ochab et~al.(2025{\natexlab{b}})Ochab, Matias, Boba \& Walkowiak}]{ochab2025styloch}
\bibinfo{author}{Ochab, J.~K.}, \bibinfo{author}{Matias, M.}, \bibinfo{author}{Boba, T.}, \& \bibinfo{author}{Walkowiak, T.} (\bibinfo{year}{2025}{\natexlab{b}}).
\newblock \bibinfo{title}{{StylOch at PAN: Gradient-boosted trees with frequency-based stylometric features}}.
\newblock In \bibinfo{editor}{J.~C. de~Albornoz}, \bibinfo{editor}{J.~Gonzalo}, \bibinfo{editor}{L.~Plaza}, \bibinfo{editor}{A.~G.~S. de~Herrera}, \bibinfo{editor}{J.~Mothe}, \bibinfo{editor}{F.~Piroi}, \bibinfo{editor}{P.~Rosso}, \bibinfo{editor}{D.~Spina}, \bibinfo{editor}{G.~Faggioli}, \& \bibinfo{editor}{N.~Ferro} (Eds.), {\it \bibinfo{booktitle}{Experimental IR Meets Multilinguality, Multimodality, and Interaction. Proceedings of the Sixteenth International Conference of the CLEF Association (CLEF 2025)}\/} Lecture Notes in Computer Science.
\newblock \bibinfo{address}{Berlin Heidelberg New York}: \bibinfo{publisher}{Springer}.
\bibitem[{Ochab \& Walkowiak(2024)}]{ochab2024}
\bibinfo{author}{Ochab, J.~K.}, \& \bibinfo{author}{Walkowiak, T.} (\bibinfo{year}{2024}).
\newblock \bibinfo{title}{Implementing interpretable models in stylometric analysis}.
\newblock In {\it \bibinfo{booktitle}{Digital {Humanities} 2024: {Conference} {Abstracts}}\/}.
\newblock \bibinfo{address}{Washington, D.C.}: \bibinfo{publisher}{George Mason University (GMU)}.
\bibitem[{Okulska et~al.(2023)Okulska, Stetsenko, Kołos, Karlińska, Głąbińska \& Nowakowski}]{okulska2023stylometrix}
\bibinfo{author}{Okulska, I.}, \bibinfo{author}{Stetsenko, D.}, \bibinfo{author}{Kołos, A.}, \bibinfo{author}{Karlińska, A.}, \bibinfo{author}{Głąbińska, K.}, \& \bibinfo{author}{Nowakowski, A.} (\bibinfo{year}{2023}).
\newblock \bibinfo{title}{Stylometrix: An open-source multilingual tool for representing stylometric vectors}.
\newblock {\it \bibinfo{journal}{arXiv preprint arXiv:2309.12810}\/}, .
\bibitem[{OpenAI(2025)}]{openai_openai_nodate}
\bibinfo{author}{OpenAI} (\bibinfo{year}{2025}).
\newblock \bibinfo{title}{{OpenAI} {API} {Reference} {Documentation}: Create chat completion}.
\newblock \URLprefix \url{https://platform.openai.com/docs/api-reference/chat/create} \bibinfo{note}{(last accessed on 2025-06-30)}.
\bibitem[{Patel et~al.(2023)Patel, Rao, Kothary, McKeown \& Callison-Burch}]{patel2023learning}
\bibinfo{author}{Patel, A.}, \bibinfo{author}{Rao, D.}, \bibinfo{author}{Kothary, A.}, \bibinfo{author}{McKeown, K.}, \& \bibinfo{author}{Callison-Burch, C.} (\bibinfo{year}{2023}).
\newblock \bibinfo{title}{Learning interpretable style embeddings via prompting llms}.
\newblock {\it \bibinfo{journal}{arXiv preprint arXiv:2305.12696}\/}, .
\bibitem[{Pedregosa et~al.(2011)Pedregosa, Varoquaux, Gramfort, Michel, Thirion, Grisel, Blondel, Prettenhofer, Weiss, Dubourg, Vanderplas, Passos, Cournapeau, Brucher, Perrot \& Duchesnay}]{scikit-learn}
\bibinfo{author}{Pedregosa, F.}, \bibinfo{author}{Varoquaux, G.}, \bibinfo{author}{Gramfort, A.}, \bibinfo{author}{Michel, V.}, \bibinfo{author}{Thirion, B.}, \bibinfo{author}{Grisel, O.}, \bibinfo{author}{Blondel, M.}, \bibinfo{author}{Prettenhofer, P.}, \bibinfo{author}{Weiss, R.}, \bibinfo{author}{Dubourg, V.}, \bibinfo{author}{Vanderplas, J.}, \bibinfo{author}{Passos, A.}, \bibinfo{author}{Cournapeau, D.}, \bibinfo{author}{Brucher, M.}, \bibinfo{author}{Perrot, M.}, \& \bibinfo{author}{Duchesnay, E.} (\bibinfo{year}{2011}).
\newblock \bibinfo{title}{Scikit-learn: Machine learning in {P}ython}.
\newblock {\it \bibinfo{journal}{Journal of Machine Learning Research}\/},  {\it \bibinfo{volume}{12}\/}, \bibinfo{pages}{2825--2830}.
\bibitem[{Przystalski et~al.(2025)Przystalski, Argasiński, Lipp \& Pacholczyk}]{przystalski_building_2025}
\bibinfo{author}{Przystalski, K.}, \bibinfo{author}{Argasiński, J.~K.}, \bibinfo{author}{Lipp, N.}, \& \bibinfo{author}{Pacholczyk, D.} (\bibinfo{year}{2025}).
\newblock {\it \bibinfo{title}{Building {Personality}-{Driven} {Language} {Models}: {How} {Neurotic} is {ChatGPT}}\/}.
\newblock Synthesis {Lectures} on {Engineering}, {Science}, and {Technology}.
\newblock \bibinfo{address}{Cham}: \bibinfo{publisher}{Springer Nature Switzerland}.
\newblock \URLprefix \url{https://link.springer.com/10.1007/978-3-031-80087-0}. \DOIprefix\doi{10.1007/978-3-031-80087-0}.
\bibitem[{Raffel et~al.(2020)Raffel, Shazeer, Roberts, Lee, Narang, Matena, Zhou, Li \& Liu}]{t5sum}
\bibinfo{author}{Raffel, C.}, \bibinfo{author}{Shazeer, N.}, \bibinfo{author}{Roberts, A.}, \bibinfo{author}{Lee, K.}, \bibinfo{author}{Narang, S.}, \bibinfo{author}{Matena, M.}, \bibinfo{author}{Zhou, Y.}, \bibinfo{author}{Li, W.}, \& \bibinfo{author}{Liu, P.~J.} (\bibinfo{year}{2020}).
\newblock \bibinfo{title}{Exploring the limits of transfer learning with a unified text-to-text transformer}.
\newblock {\it \bibinfo{journal}{Journal of machine learning research}\/},  {\it \bibinfo{volume}{21}\/}, \bibinfo{pages}{1--67}.
\bibitem[{{\v R}eh{\r u}{\v r}ek \& Sojka(2010)}]{gensim}
\bibinfo{author}{{\v R}eh{\r u}{\v r}ek, R.}, \& \bibinfo{author}{Sojka, P.} (\bibinfo{year}{2010}).
\newblock \bibinfo{title}{{Software Framework for Topic Modelling with Large Corpora}}.
\newblock In {\it \bibinfo{booktitle}{{Proceedings of the LREC 2010 Workshop on New Challenges for NLP Frameworks}}\/} (pp. \bibinfo{pages}{45--50}).
\newblock \bibinfo{address}{Valletta, Malta}: \bibinfo{publisher}{ELRA}.
\bibitem[{Sadasivan et~al.(2025)Sadasivan, Kumar, Balasubramanian, Wang \& Feizi}]{sadasivan_can_2025}
\bibinfo{author}{Sadasivan, V.~S.}, \bibinfo{author}{Kumar, A.}, \bibinfo{author}{Balasubramanian, S.}, \bibinfo{author}{Wang, W.}, \& \bibinfo{author}{Feizi, S.} (\bibinfo{year}{2025}).
\newblock \bibinfo{title}{Can {AI}-{Generated} {Text} be {Reliably} {Detected}? {Stress} {Testing} {AI} {Text} {Detectors} {Under} {Various} {Attacks}}.
\newblock {\it \bibinfo{journal}{Transactions on Machine Learning Research}\/}, . \URLprefix \url{https://openreview.net/forum?id=OOgsAZdFOt}.
\bibitem[{Sarvazyan et~al.(2023{\natexlab{a}})Sarvazyan, Gonz{'a}lez, Franco-Salvador, Rangel, Chulvi \& Rosso}]{sarvazyan_overview_2023}
\bibinfo{author}{Sarvazyan, A.~M.}, \bibinfo{author}{Gonz{'a}lez, J.~A.}, \bibinfo{author}{Franco-Salvador, M.}, \bibinfo{author}{Rangel, F.}, \bibinfo{author}{Chulvi, B.}, \& \bibinfo{author}{Rosso, P.} (\bibinfo{year}{2023}{\natexlab{a}}).
\newblock \bibinfo{title}{Overview of {AuTexTification} at {IberLEF} 2023: {Detection} and {Attribution} of {Machine}-{Generated} {Text} in {Multiple} {Domains}}.
\newblock In {\it \bibinfo{booktitle}{Procesamiento del {Lenguaje} {Natural}}\/}.
\newblock \bibinfo{address}{Jaén, Spain}.
\bibitem[{Sarvazyan et~al.(2023{\natexlab{b}})Sarvazyan, Gonz{'a}lez, Rosso \& Franco-Salvador}]{sarvazyan_supervised_2023}
\bibinfo{author}{Sarvazyan, A.~M.}, \bibinfo{author}{Gonz{'a}lez, J.~A.}, \bibinfo{author}{Rosso, P.}, \& \bibinfo{author}{Franco-Salvador, M.} (\bibinfo{year}{2023}{\natexlab{b}}).
\newblock \bibinfo{title}{Supervised {Machine}-{Generated} {Text} {Detectors}: {Family} and {Scale} {Matters}}.
\newblock In \bibinfo{editor}{A.~Arampatzis}, \bibinfo{editor}{E.~Kanoulas}, \bibinfo{editor}{T.~Tsikrika}, \bibinfo{editor}{S.~Vrochidis}, \bibinfo{editor}{A.~Giachanou}, \bibinfo{editor}{D.~Li}, \bibinfo{editor}{M.~Aliannejadi}, \bibinfo{editor}{M.~Vlachos}, \bibinfo{editor}{G.~Faggioli}, \& \bibinfo{editor}{N.~Ferro} (Eds.), {\it \bibinfo{booktitle}{Experimental {IR} {Meets} {Multilinguality}, {Multimodality}, and {Interaction}}\/} (pp. \bibinfo{pages}{121--132}).
\newblock \bibinfo{address}{Cham}: \bibinfo{publisher}{Springer Nature Switzerland}.
\newblock \DOIprefix\doi{10.1007/978-3-031-42448-9_11}.
\bibitem[{Schuster et~al.(2020)Schuster, Schuster, Shah \& Barzilay}]{schuster_limitations_2020}
\bibinfo{author}{Schuster, T.}, \bibinfo{author}{Schuster, R.}, \bibinfo{author}{Shah, D.~J.}, \& \bibinfo{author}{Barzilay, R.} (\bibinfo{year}{2020}).
\newblock \bibinfo{title}{The {Limitations} of {Stylometry} for {Detecting} {Machine}-{Generated} {Fake} {News}}.
\newblock {\it \bibinfo{journal}{Computational Linguistics}\/},  {\it \bibinfo{volume}{46}\/}, \bibinfo{pages}{499--510}. \URLprefix \url{https://doi.org/10.1162/coli_a_00380}. \DOIprefix\doi{10.1162/coli_a_00380}.
\bibitem[{Stamatatos(2009)}]{stamatatos2009survey}
\bibinfo{author}{Stamatatos, E.} (\bibinfo{year}{2009}).
\newblock \bibinfo{title}{A survey of modern authorship attribution methods}.
\newblock {\it \bibinfo{journal}{Journal of the American Society for information Science and Technology}\/},  {\it \bibinfo{volume}{60}\/}, \bibinfo{pages}{538--556}.
\bibitem[{Stiff \& Johansson(2022)}]{stiff_detecting_2022}
\bibinfo{author}{Stiff, H.}, \& \bibinfo{author}{Johansson, F.} (\bibinfo{year}{2022}).
\newblock \bibinfo{title}{Detecting computer-generated disinformation}.
\newblock {\it \bibinfo{journal}{International Journal of Data Science and Analytics}\/},  {\it \bibinfo{volume}{13}\/}, \bibinfo{pages}{363--383}. \URLprefix \url{https://doi.org/10.1007/s41060-021-00299-5}. \DOIprefix\doi{10.1007/s41060-021-00299-5}.
\bibitem[{Su et~al.(2024)Su, Wu, Zhou, Ma \& Hu}]{su_hc3_2024}
\bibinfo{author}{Su, Z.}, \bibinfo{author}{Wu, X.}, \bibinfo{author}{Zhou, W.}, \bibinfo{author}{Ma, G.}, \& \bibinfo{author}{Hu, S.} (\bibinfo{year}{2024}).
\newblock \bibinfo{title}{{HC3} {Plus}: {A} {Semantic}-{Invariant} {Human} {ChatGPT} {Comparison} {Corpus}}.
\newblock \URLprefix \url{http://arxiv.org/abs/2309.02731}. \DOIprefix\doi{10.48550/arXiv.2309.02731} \bibinfo{note}{arXiv:2309.02731 [cs]}.
\bibitem[{Sun \& Lv(2025)}]{sun2025zero}
\bibinfo{author}{Sun, J.}, \& \bibinfo{author}{Lv, Z.} (\bibinfo{year}{2025}).
\newblock \bibinfo{title}{Zero-shot detection of llm-generated text via text reorder}.
\newblock {\it \bibinfo{journal}{Neurocomputing}\/},  {\it \bibinfo{volume}{631}\/}, \bibinfo{pages}{129829}.
\bibitem[{Touvron et~al.(2023)Touvron, Lavril, Izacard et~al.}]{llama2}
\bibinfo{author}{Touvron, H.}, \bibinfo{author}{Lavril, T.}, \bibinfo{author}{Izacard, G.} et~al. (\bibinfo{year}{2023}).
\newblock \bibinfo{title}{Llama: Open and efficient foundation language models}.
\newblock \href{http://arxiv.org/abs/2302.13971}{\tt arXiv:2302.13971}.
\bibitem[{Uchendu et~al.(2021)Uchendu, Ma, Le, Zhang \& Lee}]{uchendu_turingbench_2021}
\bibinfo{author}{Uchendu, A.}, \bibinfo{author}{Ma, Z.}, \bibinfo{author}{Le, T.}, \bibinfo{author}{Zhang, R.}, \& \bibinfo{author}{Lee, D.} (\bibinfo{year}{2021}).
\newblock \bibinfo{title}{{TURINGBENCH}: {A} {Benchmark} {Environment} for {Turing} {Test} in the {Age} of {Neural} {Text} {Generation}}.
\newblock In \bibinfo{editor}{M.-F. Moens}, \bibinfo{editor}{X.~Huang}, \bibinfo{editor}{L.~Specia}, \& \bibinfo{editor}{S.~W.-t. Yih} (Eds.), {\it \bibinfo{booktitle}{Findings of the {Association} for {Computational} {Linguistics}: {EMNLP} 2021}\/} (pp. \bibinfo{pages}{2001--2016}).
\newblock \bibinfo{address}{Punta Cana, Dominican Republic}: \bibinfo{publisher}{Association for Computational Linguistics}.
\newblock \URLprefix \url{https://aclanthology.org/2021.findings-emnlp.172/}. \DOIprefix\doi{10.18653/v1/2021.findings-emnlp.172}.
\bibitem[{Vaze et~al.(2022)Vaze, Han, Vedaldi \& Zisserman}]{vaze_open-set_2022}
\bibinfo{author}{Vaze, S.}, \bibinfo{author}{Han, K.}, \bibinfo{author}{Vedaldi, A.}, \& \bibinfo{author}{Zisserman, A.} (\bibinfo{year}{2022}).
\newblock \bibinfo{title}{Open-{Set} {Recognition}: {A} {Good} {Closed}-{Set} {Classifier} is {All} {You} {Need}}.
\newblock In {\it \bibinfo{booktitle}{International {Conference} on {Learning} {Representations}}\/}.
\newblock \URLprefix \url{https://openreview.net/forum?id=5hLP5JY9S2d}.
\bibitem[{Wang et~al.(2024{\natexlab{a}})Wang, Cui, Zhang, Zhang, Wang \& Liu}]{wang_towards_2024}
\bibinfo{author}{Wang, S.}, \bibinfo{author}{Cui, S.}, \bibinfo{author}{Zhang, C.}, \bibinfo{author}{Zhang, Z.}, \bibinfo{author}{Wang, J.}, \& \bibinfo{author}{Liu, T.} (\bibinfo{year}{2024}{\natexlab{a}}).
\newblock \bibinfo{title}{Towards {Persona}-{Oriented} {LLM}-{Generated} {Text} {Detection}: {Benchmark} {Dataset} and {Method}}.
\newblock In \bibinfo{editor}{M.~Wand}, \bibinfo{editor}{K.~Malinovsk{'a}}, \bibinfo{editor}{J.~Schmidhuber}, \& \bibinfo{editor}{I.~V. Tetko} (Eds.), {\it \bibinfo{booktitle}{Artificial {Neural} {Networks} and {Machine} {Learning} – {ICANN} 2024}\/} (pp. \bibinfo{pages}{352--367}).
\newblock \bibinfo{address}{Cham}: \bibinfo{publisher}{Springer Nature Switzerland}.
\newblock \DOIprefix\doi{10.1007/978-3-031-72350-6_24}.
\bibitem[{Wang et~al.(2024{\natexlab{b}})Wang, Mansurov, Ivanov, Su, Shelmanov, Tsvigun, Whitehouse, Mohammed~Afzal, Mahmoud, Sasaki, Arnold, Aji, Habash, Gurevych \& Nakov}]{wang_m4_2024}
\bibinfo{author}{Wang, Y.}, \bibinfo{author}{Mansurov, J.}, \bibinfo{author}{Ivanov, P.}, \bibinfo{author}{Su, J.}, \bibinfo{author}{Shelmanov, A.}, \bibinfo{author}{Tsvigun, A.}, \bibinfo{author}{Whitehouse, C.}, \bibinfo{author}{Mohammed~Afzal, O.}, \bibinfo{author}{Mahmoud, T.}, \bibinfo{author}{Sasaki, T.}, \bibinfo{author}{Arnold, T.}, \bibinfo{author}{Aji, A.~F.}, \bibinfo{author}{Habash, N.}, \bibinfo{author}{Gurevych, I.}, \& \bibinfo{author}{Nakov, P.} (\bibinfo{year}{2024}{\natexlab{b}}).
\newblock \bibinfo{title}{M4: {Multi}-generator, {Multi}-domain, and {Multi}-lingual {Black}-{Box} {Machine}-{Generated} {Text} {Detection}}.
\newblock In \bibinfo{editor}{Y.~Graham}, \& \bibinfo{editor}{M.~Purver} (Eds.), {\it \bibinfo{booktitle}{Proceedings of the 18th {Conference} of the {European} {Chapter} of the {Association} for {Computational} {Linguistics} ({Volume} 1: {Long} {Papers})}\/} (pp. \bibinfo{pages}{1369--1407}).
\newblock \bibinfo{address}{St. Julian's, Malta}: \bibinfo{publisher}{Association for Computational Linguistics}.
\newblock \URLprefix \url{https://aclanthology.org/2024.eacl-long.83/}.
\bibitem[{Wu et~al.(2025)Wu, Yang, Zhan, Yuan, Chao \& Wong}]{wu_survey_2025}
\bibinfo{author}{Wu, J.}, \bibinfo{author}{Yang, S.}, \bibinfo{author}{Zhan, R.}, \bibinfo{author}{Yuan, Y.}, \bibinfo{author}{Chao, L.~S.}, \& \bibinfo{author}{Wong, D.~F.} (\bibinfo{year}{2025}).
\newblock \bibinfo{title}{A {Survey} on {LLM}-{Generated} {Text} {Detection}: {Necessity}, {Methods}, and {Future} {Directions}}.
\newblock {\it \bibinfo{journal}{Computational Linguistics}\/},  (pp. \bibinfo{pages}{1--64}). \URLprefix \url{https://doi.org/10.1162/coli_a_00549}. \DOIprefix\doi{10.1162/coli_a_00549}.
\bibitem[{Xu et~al.(2024)Xu, Zhang \& Sheng}]{xu2024freqmark}
\bibinfo{author}{Xu, Z.}, \bibinfo{author}{Zhang, K.}, \& \bibinfo{author}{Sheng, V.~S.} (\bibinfo{year}{2024}).
\newblock \bibinfo{title}{Freqmark: Frequency-based watermark for sentence-level detection of llm-generated text}.
\newblock {\it \bibinfo{journal}{arXiv preprint arXiv:2410.10876}\/}, .
\bibitem[{Yadagiri et~al.(2024)Yadagiri, Kalita, Ranjan, Bostan, Toppo \& Pakray}]{yadagiri_team_2024}
\bibinfo{author}{Yadagiri, A.}, \bibinfo{author}{Kalita, D.}, \bibinfo{author}{Ranjan, A.}, \bibinfo{author}{Bostan, A.~K.}, \bibinfo{author}{Toppo, P.}, \& \bibinfo{author}{Pakray, P.} (\bibinfo{year}{2024}).
\newblock \bibinfo{title}{Team cnlp-nits-pp at {PAN}: {Leveraging} {BERT} for {Accurate} {Authorship} {Verification}: {A} {Novel} {Approach} to {Textual} {Attribution}}.
\newblock In \bibinfo{editor}{G.~Faggioli}, \bibinfo{editor}{N.~Ferro}, \bibinfo{editor}{P.~Galušč{'a}kov{'a}}, \& \bibinfo{editor}{A.~G.~S. Herrera} (Eds.), {\it \bibinfo{booktitle}{Working {Notes} {Papers} of the {CLEF} 2024 {Evaluation} {Labs}}\/} (pp. \bibinfo{pages}{2976--2987}).
\newblock \bibinfo{publisher}{CEUR-WS.org}.
\newblock \URLprefix \url{http://ceur-ws.org/Vol-3740/paper-290.pdf}.
\bibitem[{Yao et~al.(2024)Yao, Duan, Xu, Cai, Sun \& Zhang}]{yao2024survey}
\bibinfo{author}{Yao, Y.}, \bibinfo{author}{Duan, J.}, \bibinfo{author}{Xu, K.}, \bibinfo{author}{Cai, Y.}, \bibinfo{author}{Sun, Z.}, \& \bibinfo{author}{Zhang, Y.} (\bibinfo{year}{2024}).
\newblock \bibinfo{title}{A survey on large language model (llm) security and privacy: The good, the bad, and the ugly}.
\newblock {\it \bibinfo{journal}{High-Confidence Computing}\/},  (p. \bibinfo{pages}{100211}).
\bibitem[{Yu et~al.(2025)Yu, Chen, Feng \& Xia}]{yu_cheat_2025}
\bibinfo{author}{Yu, P.}, \bibinfo{author}{Chen, J.}, \bibinfo{author}{Feng, X.}, \& \bibinfo{author}{Xia, Z.} (\bibinfo{year}{2025}).
\newblock \bibinfo{title}{{CHEAT}: {A} {Large}-scale {Dataset} for {Detecting} {CHatGPT}-{writtEn} {AbsTracts}}.
\newblock {\it \bibinfo{journal}{IEEE Transactions on Big Data}\/},  (pp. \bibinfo{pages}{1--9}). \URLprefix \url{https://ieeexplore.ieee.org/abstract/document/10858415}. \DOIprefix\doi{10.1109/TBDATA.2025.3536929}.
\bibitem[{Zellers et~al.(2019)Zellers, Holtzman, Rashkin, Bisk, Farhadi, Roesner \& Choi}]{zellers_defending_2019}
\bibinfo{author}{Zellers, R.}, \bibinfo{author}{Holtzman, A.}, \bibinfo{author}{Rashkin, H.}, \bibinfo{author}{Bisk, Y.}, \bibinfo{author}{Farhadi, A.}, \bibinfo{author}{Roesner, F.}, \& \bibinfo{author}{Choi, Y.} (\bibinfo{year}{2019}).
\newblock \bibinfo{title}{Defending {Against} {Neural} {Fake} {News}}.
\newblock In \bibinfo{editor}{H.~Wallach}, \bibinfo{editor}{H.~Larochelle}, \bibinfo{editor}{A.~Beygelzimer}, \bibinfo{editor}{F.~d. Alché-Buc}, \bibinfo{editor}{E.~Fox}, \& \bibinfo{editor}{R.~Garnett} (Eds.), {\it \bibinfo{booktitle}{Advances in {Neural} {Information} {Processing} {Systems}}\/}.
\newblock \bibinfo{publisher}{Curran Associates, Inc.} volume~\bibinfo{volume}{32}.
\newblock \URLprefix \url{https://proceedings.neurips.cc/paper_files/paper/2019/file/3e9f0fc9b2f89e043bc6233994dfcf76-Paper.pdf}.
\bibitem[{Zhao et~al.(2023)Zhao, Zhou, Li, Tang, Wang, Hou, Min, Zhang, Zhang, Dong et~al.}]{zhao2023survey}
\bibinfo{author}{Zhao, W.~X.}, \bibinfo{author}{Zhou, K.}, \bibinfo{author}{Li, J.}, \bibinfo{author}{Tang, T.}, \bibinfo{author}{Wang, X.}, \bibinfo{author}{Hou, Y.}, \bibinfo{author}{Min, Y.}, \bibinfo{author}{Zhang, B.}, \bibinfo{author}{Zhang, J.}, \bibinfo{author}{Dong, Z.} et~al. (\bibinfo{year}{2023}).
\newblock \bibinfo{title}{A survey of large language models}.
\newblock {\it \bibinfo{journal}{arXiv preprint arXiv:2303.18223}\/}, .

\end{thebibliography}







\end{document}